\documentclass[runningheads]{llncs}

\usepackage{eccv}
\usepackage{eccvabbrv}

\usepackage{graphicx}
\usepackage{booktabs}

\usepackage[accsupp]{axessibility}  % Improves PDF readability for those with disabilities.

\usepackage[pagebackref,breaklinks,colorlinks,citecolor=eccvblue]{hyperref}
\usepackage{multirow}
\usepackage{subcaption}
\usepackage{graphicx} % placeholder
\usepackage{mwe}  % example-image 포함 placeholder
\usepackage{wrapfig}
\usepackage{lipsum}
\usepackage{pifont} % check marks in ablation table
\usepackage[table]{xcolor}
\newcommand{\hlrow}[1]{\rowcolor{#1}}
\usepackage{bbm}
\usepackage{makecell}
\usepackage{enumitem}

\newcommand{\OursVFI}{\textbf{TETO-VFI}}

\newcommand{\Ours}{\textbf{TETO}}
\newcommand{\cmark}{\ding{51}}                 % ✓ (black)
\newcommand{\xmark}{\textcolor{gray!60}{\ding{55}}} % ✗ (gray)
\definecolor{myblue}{RGB}{220,230,255}
\definecolor{mygreen}{RGB}{0,143,0}

\begin{document}

\title{TETO: Tracking Events with Teacher Observation for Motion Estimation and Frame Interpolation} 
% \maketitle
\titlerunning{TETO: Tracking Events with Teacher Observation}

\author{Jini Yang$^{\ast}$\inst{1} \and Eunbeen Hong$^{\ast}$\inst{1} \and Soowon Son\inst{1} \and Hyunkoo Lee\inst{1} \and Sunghwan Hong\inst{2} \and Sunok Kim\inst{3} \and Seungryong Kim\inst{1}}

\authorrunning{Yang et al.}

\institute{KAIST AI \and
ETH Zurich \and Korea Aerospace University\\[1em]
Project page: \url{https://cvlab-kaist.github.io/TETO}}

\maketitle
\begingroup
\renewcommand{\thefootnote}{}
\footnotetext{\scriptsize $\ast$: Equal contribution}
\endgroup

\begin{abstract}
Event cameras capture per-pixel brightness changes with microsecond resolution, offering continuous motion information lost between RGB frames. However, existing event-based motion estimators depend on large-scale synthetic data that often suffers from a significant sim-to-real gap. We propose \textbf{TETO} (\textbf{T}racking \textbf{E}vents with \textbf{T}eacher \textbf{O}bservation), a teacher-student framework that learns event motion estimation from only $\sim$25 minutes of unannotated real-world recordings through knowledge distillation from a pretrained RGB tracker. Our motion-aware data curation and query sampling strategy maximizes learning from limited data by disentangling object motion from dominant ego-motion. The resulting estimator jointly predicts point trajectories and dense optical flow, which we leverage as explicit motion priors to condition a pretrained video diffusion transformer for frame interpolation. We achieve state-of-the-art point tracking on EVIMO2 and optical flow on DSEC using orders of magnitude less training data, and demonstrate that accurate motion estimation translates directly to superior frame interpolation quality on BS-ERGB and HQ-EVFI.
\end{abstract}
\section{Introduction}
\label{sec:intro}
Motion estimation, represented as point tracking~\cite{karaev2024cotracker, cho2024local, karaev2025cotracker3, harley2025alltracker, doersch2023tapir, doersch2024bootstap} and optical flow~\cite{wang2024sea, teed2020raft, harley2025alltracker}, is fundamental to computer vision, with applications in robotics~\cite{vecerik2024robotap}, autonomous driving~\cite{luo2021exploring}, 3D/4D reconstruction~\cite{hong2024unifying, an2025c3g, han2025d}, and video generation~\cite{geng2025motion, jeong2025track4gen, briedis2025controllable, wang2025ati, chu2025wan}. Recent RGB-based methods have achieved impressive performance~\cite{doersch2024bootstap, karaev2024cotracker, karaev2025cotracker3, harley2025alltracker}, but these methods rely on discrete frame sampling and inevitably fail under fast motion or challenging lighting, where inter-frame information is lost.

\begin{figure}[t!]
    \centering
    \includegraphics[width=0.95\linewidth]{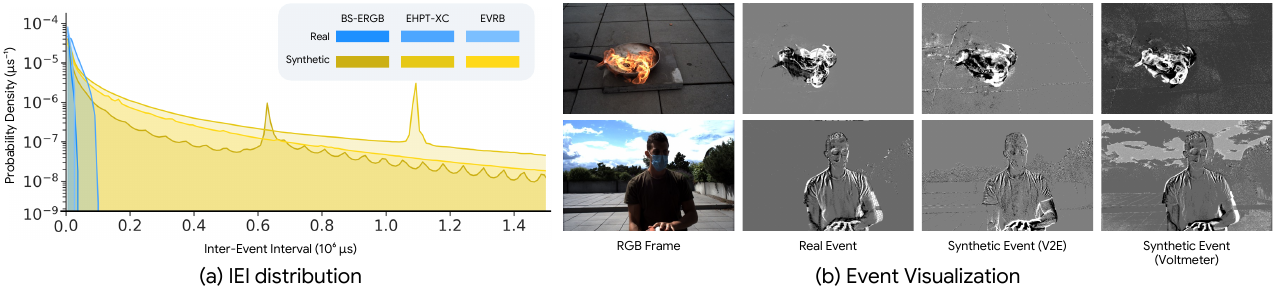}
    \caption{\textbf{Real event vs. synthetic event analysis.} (a) Inter-Event Interval (IEI) distributions show that real events concentrate in short intervals with rapid decay, while synthetic events exhibit long tails and periodic artifacts. (b) The appearance of real and synthetic events differs substantially, with synthetic events from V2E~\cite{hu2021v2e} and Voltmeter~\cite{lin2022dvs} exhibiting artifacts not present in real-world recordings.
}
    \vspace{-10pt}
    \label{fig:event_sim2real}
\end{figure}

Event cameras address this limitation by recording per-pixel brightness changes asynchronously with microsecond temporal resolution, providing continuous motion measurements between frames~\cite{gallego2020event}. This unique capability makes event cameras particularly promising for motion estimation in dynamic scenes involving fast-moving objects. Such reliable blind-time motion estimates not only enable robust tracking under challenging conditions~\cite{shen2025blinktrack, hamann2025etap}, but also serve as explicit motion representations for downstream tasks such as video frame interpolation~\cite{liu2025timetracker}, where the quality of motion prediction directly determines synthesis fidelity.
\begin{wrapfigure}{r}{0.5\linewidth}
    \vspace{-20pt}
    \centering
    \includegraphics[width=\linewidth]{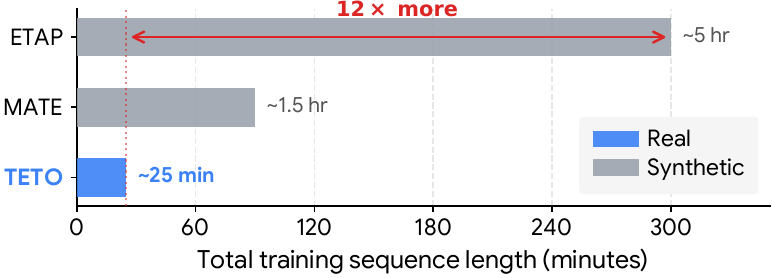}
    \vspace{-15pt}
    \caption{\textbf{Training data scale of event-based trackers.}}
    \vspace{-20pt}
    \label{fig:event_data}
\end{wrapfigure}
While recent event-based motion estimation methods~\cite{hamann2025etap, han2025mate} leverage the advantage of event cameras and achieve strong performance under fast motion, they rely heavily on large-scale synthetic data and inevitably lack generalizability. Training only on synthetic data does not adequately reflect the diverse and complex scenarios encountered in real-world dynamic scenes, including object interactions and frequent occlusions. As illustrated in Figure~\ref{fig:event_data}, these approaches depend on extensive synthetic generation pipelines, ETAP~\cite{hamann2025etap} trained on approximately 5 hours of sequences and MATE~\cite{han2025mate} on 1.5 hours, incurring substantial computational costs for rendering~\cite{greff2022kubric, zheng2023pointodyssey} RGB frames before processed with RGB frame interpolation models~\cite{reda2022film, huang2022real}, and event synthesized with those upsampled frames~\cite{rebecq2018esim, lin2022dvs, hu2021v2e}. In addition, synthesizing events from discretely interpolated RGB frames oversimplifies continuous motion dynamics. Figure~\ref{fig:event_sim2real} shows Inter-Event Interval (IEI) distribution difference of real and synthetic event. Notably, synthetic events exhibit periodic artifacts induced by interpolation, leading to a sim-to-real gap and limited generalization to complex real-world motion.
We address these challenges by asking: \textbf{{How can we achieve event motion estimation with only a small amount of unannotated real-world data?}} The key obstacle is the absence of motion annotations for real event sequences. We overcome this through knowledge distillation from a pretrained RGB tracker, which generates pseudo trajectory and optical flow labels on real event-RGB paired sequences without manual annotation.

Here, we propose \textbf{TETO} (\textbf{T}racking \textbf{E}vents with \textbf{T}eacher \textbf{O}bservation), a teacher-student framework for learning event motion estimation from limited real-world data. We carefully curate $\sim$25 minutes of real event-RGB recordings that meet strict criteria for sensor alignment, motion diversity, and spatial resolution. To maximize learning from this limited data, we decompose the teacher's optical flow to disentangle object motion from dominant ego-motion and oversample query points from dynamic regions, ensuring robust correspondences for both global and local motion. For the student architecture, we adapt AllTracker~\cite{harley2025alltracker} with a concentration network that converts event stacks into a 3-channel representation, enabling joint prediction of point trajectories and dense optical flow from events without architectural modification.

Since our motion estimator jointly outputs dense optical flow and point trajectories, the resulting blind-time motion estimates naturally extend to video frame interpolation. We condition a pretrained video diffusion transformer~\cite{wan2025wan} with complementary motion signals. Specifically, optical flow from TETO warps boundary frame latents for coarse spatial alignment, point trajectories supervise attention maps as geometric guidance~\cite{jeong2025track4gen, nam2025emergent, son2025repurposing}, and an event motion mask derived from raw event accumulations directs generation capacity toward dynamic regions. These signals tightly constrain the synthesis process, and evaluation confirms that generalizable motion estimation translates to superior frame quality.

We evaluate TETO across point tracking, optical flow, and video frame interpolation on multiple real-world benchmarks. Our method achieves state-of-the-art point tracking on EVIMO2~\cite{burner2022evimo2} despite using orders of magnitude less data than synthetic-trained approaches, and outperforms all unsupervised and self-supervised optical flow methods on DSEC~\cite{gehrig2021dsec} with in-domain fine-tuning while remaining competitive in the zero-shot setting. For video frame interpolation, we achieve the best perceptual quality on BS-ERGB~\cite{tulyakov2022time} and HQ-EVFI~\cite{ma2024timelens}.

Our contributions are as follows:
\begin{itemize}
\renewcommand{\labelitemi}{$\bullet$}
\item We propose an event-based motion estimator that learns
dynamic motion estimation from only $\sim$25 minutes of
unannotated real-world recordings, with motion-aware data curation
and query sampling.
\item We present a video frame interpolation framework that
conditions a pretrained video diffusion transformer with dense
optical flow and point trajectories from our motion estimator,
along with event-derived motion masks, as strong motion priors
for accurate dynamic scene synthesis.
\item We achieve state-of-the-art point tracking on EVIMO2 and
optical flow on DSEC, demonstrating that accurate dynamic motion
estimation enables high-quality frame generation on BS-ERGB and
HQ-EVFI.
\end{itemize}
\section{Related Work}
\label{sec:rel_works}

\subsection{Event-based motion estimation}
\label{sec:rel_motion}
% optical flow, point tracking
\subsubsection{Event point tracking.}
Point tracking estimates long-range correspondences of arbitrary query points across video frames, enabling detailed analysis of object motion in dynamic scenes. RGB-based trackers have achieved strong performance by training on large-scale synthetic datasets such as Kubric~\cite{greff2022kubric}. However, recent data-centric approaches demonstrate that real motion data can be more effective than scaling synthetic generation. CoTracker3~\cite{karaev2025cotracker3} leverages pseudo-labels from pretrained teachers to fine-tune on real videos, and AnthroTAP~\cite{kim2025anthrotap} constructs a curated dataset capturing complex human motion, both substantially improving tracking robustness.

This shift toward prioritizing data quality over dataset scale has not yet reached the event domain. ETAP~\cite{hamann2025etap} and MATE~\cite{han2025mate} extend the Tracking Any Point paradigm to event cameras but train entirely on synthetic datasets such as EventKubric and Ev-PointOdyssey. Creating these datasets requires expensive pipelines combining scene rendering, RGB video interpolation, and event simulation. Such pipelines are not only computationally costly to scale but also inevitably introduce unrealistic motion patterns and noise statistics, leading to a noticeable sim-to-real gap when applied to real dynamic scenes.

\subsubsection{Event optical flow.}
Optical flow estimates dense per-pixel motion between consecutive frames and provides spatial alignment cues essential for downstreamtasks such as video frame interpolation. Early event-based approaches such as EV-FlowNet~\cite{zhu2019evflownet}adopt self-supervised training without requiring ground-truth annotations but often struggle to generalize beyond the training distribution. Supervised methods including E-RAFT~\cite{gehrig2021raft} and TMA~\cite{liu2023tma} significantly improve accuracy on labeled datasets such as DSEC~\cite{gehrig2021dsec}, and more recent work extends flow estimation to continuous-time formulations~\cite{gehrig2024dense} and cross-modal fusion between events and frames~\cite{zhou2025bridge}. Despite these advances, evaluation protocols remain heavily biased
toward driving scenarios dominated by ego-motion, leaving dynamic
object motion relatively underexplored.

\subsection{Event-based video frame interpolation}
\label{sec:rel_vfi}

Event-based video frame interpolation (EVFI) synthesizes intermediate frames by exploiting the high temporal resolution of event cameras between sparse RGB observations. TimeLens~\cite{tulyakov2021time} introduced a flow-based warping framework driven by event representations, with TimeLens++~\cite{tulyakov2022time} and TimeLens-XL~\cite{ma2024timelens} extending the formulation to handle non-linear motion and larger displacements. CBMNet~\cite{kim2023event} estimates bidirectional motion fields jointly across frame and event modalities, and TimeTracker~\cite{liu2025timetracker} connects point tracking to EVFI by deriving coarse flow from event-based trajectories before refinement. While these warping-based approaches produce strong reconstruction, their accuracy remains sensitive to flow estimation errors near occlusions and complex motion boundaries.

More recent work incorporates diffusion-based generation to address these limitations. RE-VDM~\cite{chen2025revdm} conditions a video diffusion model directly on event representations to improve cross-camera generalization, and EventDiff~\cite{zheng2025eventdiff} introduces a hybrid event-frame autoencoder with latent-space denoising. These approaches advance modality integration and generation quality, but rely on implicit motion cues encoded in event representations rather than explicit motion estimation.
\section{Event-based motion estimation}
\label{sec:tracking}
We train an event motion estimator only on 25 minutes of paired event-RGB recordings without motion annotations. Given an event stream and query points, the model jointly predicts point trajectories with visibility estimates and dense optical flow from a single forward pass, capturing both long-range sparse tracking and dense local displacement.

\subsection{Preliminaries and problem formulation}

Event cameras asynchronously output events $e_i = (\mathbf{x}_i, t_i, p_i)$, 
where $\mathbf{x}_i \in \mathbb{Z}^2$ is the pixel location, $t_i$ is the 
timestamp, and $p_i \in \{-1, +1\}$ is the polarity of the brightness 
change. Unlike spatially dense RGB frames, events are spatially sparse 
and temporally continuous, requiring an intermediate representation for 
compatibility with frame-based architectures. At each timestamp $t$, we 
collect the most recent $N$ events preceding $t$, where $N$ denotes the total event count per timestamp. From these, we construct 
$B$ stacks with exponentially increasing event counts, where the $b$-th 
stack ($b = 1, \ldots, B$) contains the most recent $N / 2^{B-b}$ events. 
Each stack is converted into a single-bin image by accumulating event 
polarities at each pixel location, forming the event stack 
$E_t \in \mathbb{R}^{H \times W \times B}$ at each timestamp $t$. We obtain event stacks $\{E_t\}_{t=1}^{T}$ and corresponding 
RGB images $\{I_t\}_{t=1}^{T}$ with $I_t \in \mathbb{R}^{H \times W \times 3}$. 

For motion estimation, given a set of query points $\mathbf{q} \in \mathbb{R}^{N_q \times 3}$ specifying 
pixel coordinates and query timestamps, our model predicts point 
trajectories $\mathcal{T} \in \mathbb{R}^{N_q \times T \times 2}$ with per-point 
visibility $v \in \{0,1\}^{N_q \times T}$ and dense optical flow 
$\mathcal{F} \in \mathbb{R}^{H \times W \times 2}$ at each timestamp. Note that RGB frames are used only for pseudo-label generation and are not required at inference.

\subsection{Real-world data curation with pseudo-labels}
\label{sec:tracking-data}

Our goal is to train an effective motion estimator from limited real paired RGB-event data. Since real-world event-RGB paired datasets remain scarce in both scale and diversity, we address this through careful dataset curation and motion-aware query sampling, detailed in the following.

\begin{table}[t]
\centering
\caption{\textbf{Public event-RGB paired datasets.} (E: Event, R: RGB, G:Grayscale.) Align indicates beamsplitter-based spatial alignment between event and frame sensors. Challenge denotes the dominant difficulty of each dataset. Both BS-ERGB and ERF-X170FPS capture extreme object motion, but the higher frame rate of ERF-X170FPS (170 vs.\ 28 FPS) allows the RGB teacher to produce stable pseudo-label trajectories. Blue rows denote our training sets.}
\resizebox{0.9\linewidth}{!}
{
\begin{tabular}{l|ccccccc}
\toprule
\textbf{Dataset} & Align & Modality & Motion dyn & Scene & Resolution & Challenge & FPS \\
\midrule
E-POSE~\cite{hay2025pose} & \xmark & E+R & Cam & Indoor & 346$\times$260  & - & -  \\
M3ED~\cite{chaney2023m3ed} & \xmark & E+R & Cam & Indoor/Outdoor & 1280$\times$720  & - & 20  \\
MTevent~\cite{awasthi2025mtevent} & \xmark & E+R & Obj+Cam & Indoor & 640$\times$480 & - & 25, 100  \\
HS-ERGB~\cite{tulyakov2021time} & \xmark & E+R & Obj+Cam & Indoor/Outdoor & 1280$\times$720  & Non-linear Motion & 240  \\
RELI11D~\cite{yan2024reli11d} & \xmark & E+R & Obj+Cam & Indoor/Outdoor & 1280$\times$800  & Complex Motion & 60  \\
DSEC~\cite{gehrig2021dsec} & \xmark & E+R & Cam & Outdoor & 640$\times$480 & Low Light & 20 \\
\midrule
HQF~\cite{stoffregen2020hqg} & \cmark & E+G & Cam & Indoor/Outdoor & 240$\times$180  & - & -  \\
EVRB~\cite{kim2024cmta} & \cmark & E+R & Cam & Outdoor & 960$\times$640 & Blur & - \\
BS-ERGB~\cite{tulyakov2022time} & \cmark & E+R & Obj+Cam & Indoor/Outdoor & 970$\times$625  & Extreme Motion  & 28  \\
\hlrow{myblue}
ERF-X170FPS~\cite{kim2023event} & \cmark & E+R & Obj+Cam & Indoor/Outdoor & 1280$\times$720  & Extreme Motion & 170  \\
\hlrow{myblue}
EHPT-XC~\cite{cho2024benchmark} & \cmark & E+R & Obj+Cam & Indoor/Outdoor & 1280$\times$720 & - & - \\
\bottomrule
\end{tabular}
}
\label{table:data_curation}
\end{table}

\subsubsection{Dataset selection.}
Training event-based motion estimators with limited real-world data requires carefully curated event-RGB paired sequences. We identify the following criteria for effective training data: (1) real event sensor recordings, (2) diverse motion dynamics encompassing both camera and object motion, (3) sufficient spatial resolution for fine-grained tracking, (4) a beamsplitter or equivalent setup ensuring precise spatial alignment, (5) motion complexity within the operating range of the RGB teacher, and (6) a mix of indoor and outdoor environments. As summarized in Table~\ref{table:data_curation}, among all surveyed public datasets, only EHPT-XC and ERF-X170FPS satisfy all six criteria simultaneously. EHPT-XC contains complex human motion with flexible body deformations and occlusions from limb interactions. ERF-X170FPS captures highly non-linear motion dynamics while its high frame rate (170 FPS) enables the RGB teacher to produce longer, more stable pseudo-label tracks.

\begin{wrapfigure}{r}{0.5\linewidth}
    \centering
    \vspace{-30pt}
    \includegraphics[width=\linewidth]{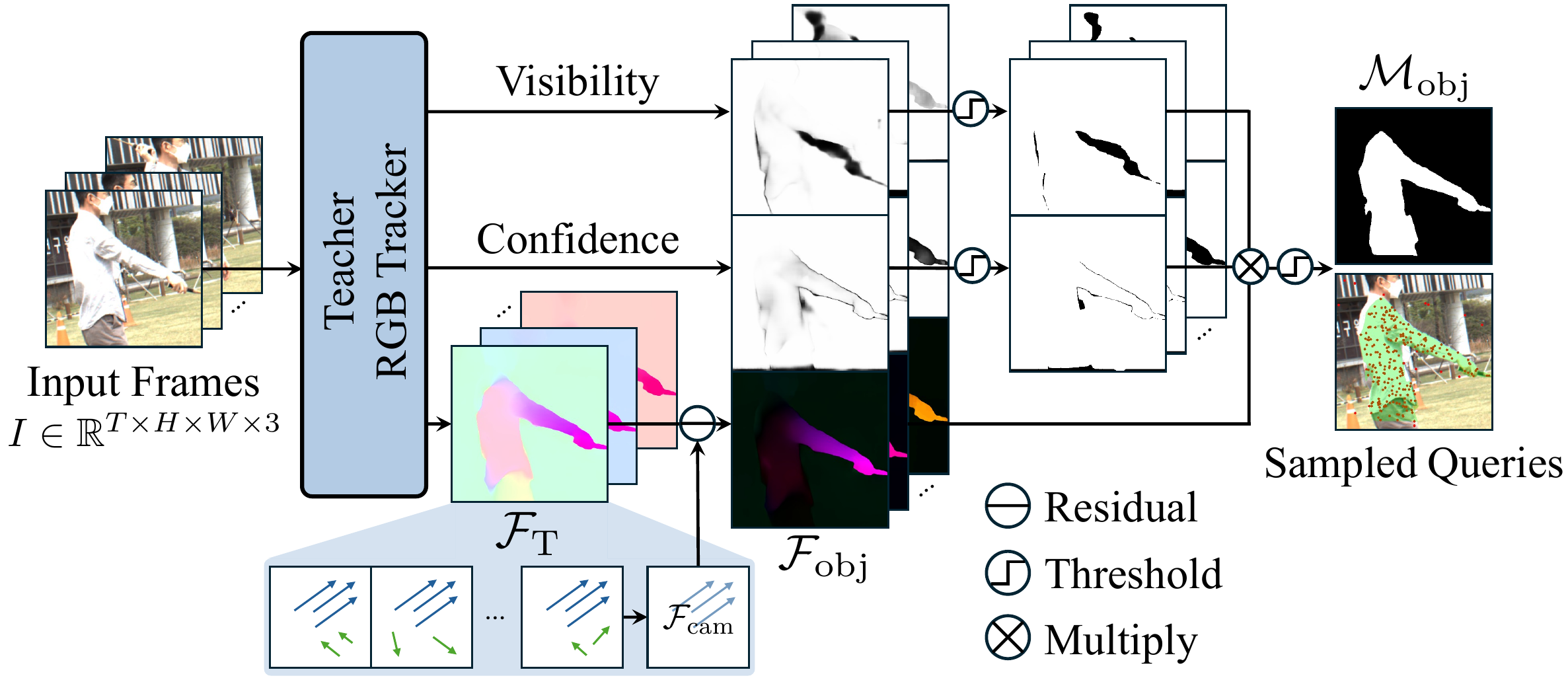}
    \caption{\textbf{Object motion query sampling.} Given teacher-predicted optical flow, 
we estimate a global affine model via RANSAC and compute residual flow to identify 
independently moving regions. Queries are oversampled from these regions to prevent 
bias toward dominant ego-motion patterns.}
    \vspace{-30pt}
    \label{fig:object-mask}
\end{wrapfigure}
\subsubsection{Motion-aware pseudo-label generation.}
We generate pseudo-labels using a pretrained RGB tracker~\cite{harley2025alltracker} that jointly predicts trajectories $\mathcal{T}_T$ and dense optical flow $\mathcal{F}_T$. A key challenge with limited data is motion distribution imbalance, where camera ego-motion dominates as global scene motion triggers significantly more events than localized object movement. Naively sampling query points leads to overfitting to this dominant pattern.

To address this, we decompose $\mathcal{F}_T$ to isolate object-specific movement. We estimate a global affine model $\mathbf{A}$ via RANSAC and derive a binary object motion mask from the residual flow magnitude at each pixel location $\mathbf{x}$:
\begin{equation}
\mathcal{M}_{\mathrm{obj}}(\mathbf{x}) = \mathbbm{1}\!\left[\,\|\mathcal{F}_T(\mathbf{x}) - \mathbf{A}[\mathbf{x}, 1]^\top\| > \tau_\text{th}\,\right],
\end{equation}
where $\tau_\text{th}$ is determined by the median absolute deviation of the 
residual flow. As shown in Figure~\ref{fig:object-mask}, we sample the majority of $N_q$ query points from $\mathcal{M}_{\mathrm{obj}}$ to ensure sufficient exposure to dynamic motion patterns while retaining background queries for global motion context. Further sampling details are in the Appendix.
\begin{figure*}[t!]
    \centering
    \includegraphics[width=1.00\textwidth]{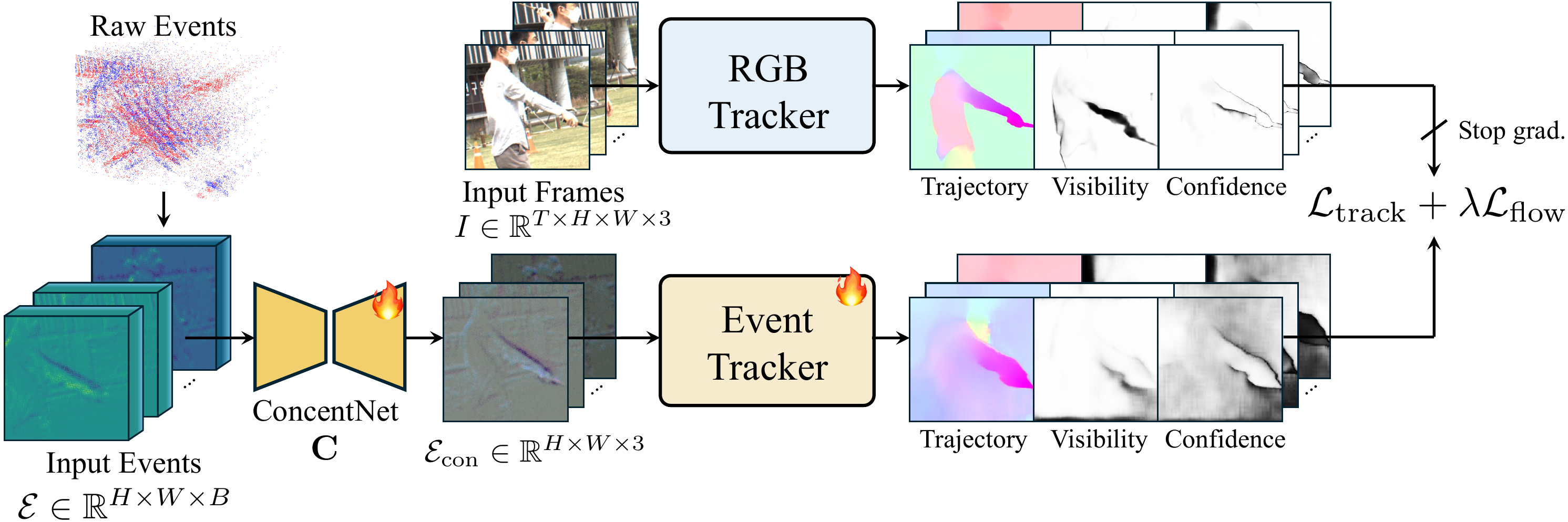}
    % \vspace{-10pt}
    \caption{\textbf{Overall architecture.} The concentration network aggregates multi-scale events into a 3-channel representation. Pseudo-labels for tracking and optical flow are provided by the teacher tracker.}
    \vspace{-10pt}
    \label{fig:track_arch}
\end{figure*}

\subsection{Distillation from pretrained RGB tracker}
\label{sec:tracking_tracker}

Rather than training an event tracker from scratch, we transfer the correspondence matching ability of a pretrained RGB tracker to the event domain through fine-tuning with pseudo-label supervision. As illustrated in Figure~\ref{fig:track_arch}, the concentration network first converts multi-scale event stacks into a 3-channel representation compatible with the RGB tracker encoder, and the downstream matching modules jointly predict point trajectories and dense optical flow. The entire pipeline is supervised by the teacher's pseudo-labels, enabling the student to inherit correspondence matching ability while learning to exploit temporal motion cues that only the event modality can provide, such as continuous brightness changes under low light or extremely fast motion.

\subsubsection{Architecture.}
Our event motion estimator is built on the AllTracker architecture~\cite{harley2025alltracker}, which jointly outputs dense optical flow and sparse point trajectories. To operate the pretrained RGB tracker on events without architectural modification, a concentration network $\mathcal{C}$, implemented as a U-Net, aggregates multi-scale temporal information~\cite{nam2022stereo} in stacked events $E_t \in \mathbb{R}^{H \times W \times B}$ into a 3-channel representation $\hat{E}_t = \mathcal{C}(E_t) \in \mathbb{R}^{H \times W \times 3}$, compressing $B$ temporal bins into three channels compatible with the pretrained RGB encoder. A ConvNeXt-based encoder extracts low-resolution feature maps at $1/8$ spatial stride from each $\hat{E}_t$. The query frame features are tiled across all timesteps and cross-correlated with target features at multiple scales to build correlation volume pyramids. A recurrent module then refines correspondence estimates over $K$ iterations, interleaving 2D convolutional blocks for spatial propagation with pixel-aligned temporal attention for multi-frame reasoning. The final estimates are upsampled to full resolution via pixel shuffle, producing dense optical flow, point trajectories, visibility, and confidence. The entire pipeline is fine-tuned end-to-end with pseudo-label supervision.

\subsubsection{Training objectives.}
The student event motion estimator is supervised with both trajectory and flow objectives. At the $k$-th refinement iteration, the student predicts trajectories $\mathcal{T}^k$ and dense optical flow $\mathcal{F}^k$. The trajectory loss $\mathcal{L}_{\text{track}}$ uses $\ell_1$ distance between estimated and pseudo-label trajectories, weighted by teacher-derived visibility $v_T$ and confidence $c_T$:

\begin{equation}
\mathcal{L}_{\text{track}} = \alpha \sum_{k=1}^{K} \gamma^{K-k} \left( \frac{\mathbbm{1}_{\text{occ}}}{5} + \mathbbm{1}_{\text{vis}} \right) \| \mathcal{T}^k- \mathcal{T}_T  \|_1 ,
\end{equation}
where $\mathbbm{1}_{\text{vis}}$ selects points that the teacher marks as visible with high confidence, $\mathbbm{1}_{\text{occ}} = 1 - \mathbbm{1}_{\text{vis}}$ denotes the remaining occluded or uncertain points, $\gamma$ is the iteration decay factor, and $\alpha$ is the trajectory loss weight.
The flow consistency loss $\mathcal{L}_{\text{flow}}$ supervises the dense correspondence field:

\begin{equation}
\mathcal{L}_{\text{flow}} = \sum_{k=1}^{K} \gamma^{K-k} \| \mathcal{F}^k - \mathcal{F}_T \|_1.
\end{equation}
Flow supervision enforces long-range dense correspondence beyond sparse query points and fully utilizes the teacher's motion knowledge at the pixel level. The total objective is $\mathcal{L} = \mathcal{L}_{\text{track}} + \lambda \mathcal{L}_{\text{flow}}$, where $\lambda$ controls the relative weight of flow supervision.

\section{Motion-aware video frame interpolation}
\label{sec:interpolation}

Given boundary frames $I_0$ and $I_1$ with an event stream between them, we synthesize intermediate frames $I_t$ at arbitrary timestamps. Accurately capturing dynamic motion during blind time is the central challenge for this task. We condition a pre-trained video diffusion transformer with three complementary signals from the estimator, where optical flow provides coarse spatial alignment, point trajectories guide fine-grained correspondences through attention, and an event motion mask $\mathcal{M}_{\mathrm{event}}$ identifies dynamic regions.

\subsection{Preliminaries}
Recent video diffusion models operate on latent video representations with a pretrained VAE. Given an input video $V \in \mathbb{R}^{(1+T)\times H \times W \times 3}$,
an encoder maps it to a latent representation
$z \in
\mathbb{R}^{(1+\frac{T}{f_t})\times \frac{H}{f_s}\times \frac{W}{f_s}\times C}$,
where $f_t$ and $f_s$ denote temporal and spatial compression ratios.
These models are often trained using flow-based generative frameworks such as rectified flow. Given a noise sample $\epsilon \sim \mathcal{N}(0, I)$ and a clean latent $z$, rectified flow defines a linear path $z_{\tau} = (1-\tau)\epsilon + \tau z$ where $\tau \sim \mathcal{U}(0,1)$.
The model 
$v_\theta(x_{\tau}, \tau)$ is trained to predict the 
velocity field via
\begin{equation}
\mathcal{L}_\text{rf} = \mathbb{E}\left[\|v_\theta(x_{\tau}, 
\tau) - (z - \epsilon)\|^2\right].
\end{equation}

The denoising network is implemented as a diffusion transformer (DiT)
which models spatial-temporal dependencies using self-attention.
Given query, key, and value features
$Q$, $K$, $V$ are obtained via linear projections, and attention is
computed as

\begin{equation}
\text{Attn}(Q,K,V) =
\text{Softmax}\left(\frac{QK^\top}{\sqrt{d}}\right)V ,
\end{equation}
where $d$ denotes the dimensionality of the key features.
\begin{figure*}[t!]
    \centering
    \includegraphics[width=1.00\textwidth]{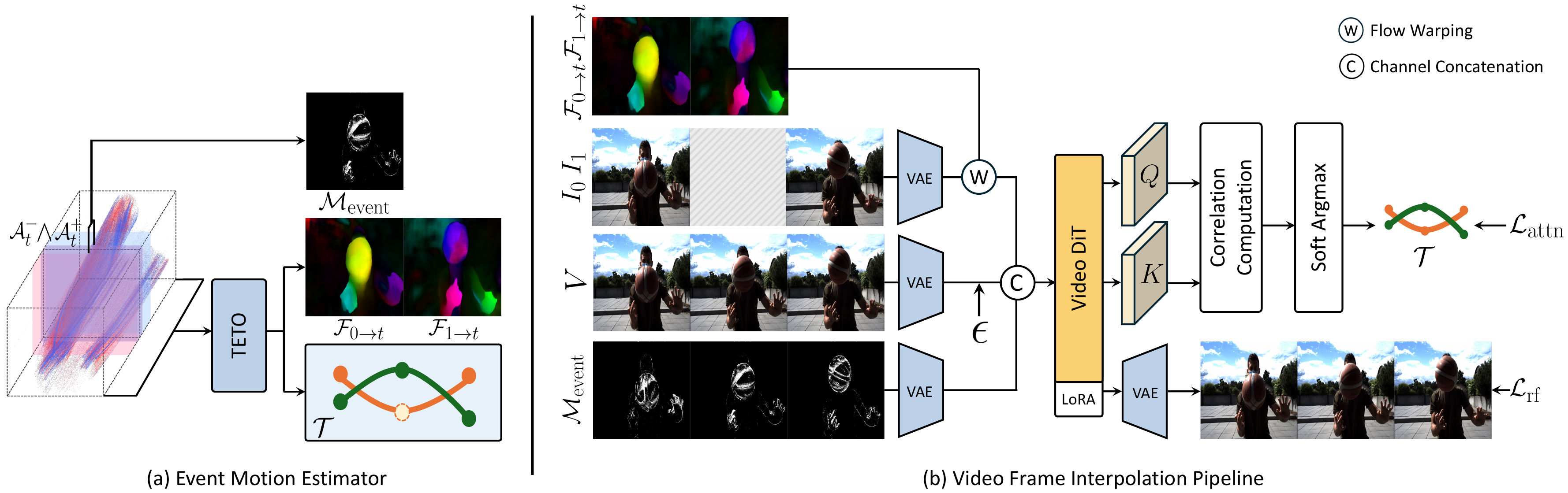}
    \caption{\textbf{Video frame interpolation architecture.} (a) Motion signals from TETO and event motion mask construction. (b) Overall pipeline with latent-space flow warping and trajectory-guided attention supervision.}
    \vspace{-10pt}
    \label{fig:interp_arch}
\end{figure*}

\subsection{Motion-conditioned frame interpolation}

Our frame interpolation pipeline conditions a pre-trained video diffusion transformer with three complementary motion signals, all derived from TETO and raw events without any event-specific learned components in the generation model as illustrated in Figure~\ref{fig:interp_arch}. Given boundary frames $I_0$ and $I_1$ with an event stream between them, TETO estimates bidirectional optical flow and point trajectories at each target timestamp $t$, while the event motion mask is constructed directly from raw event accumulations. These signals are injected at two levels of the diffusion process: optical flow and the event motion mask operate at the latent input level to provide spatial alignment and dynamic region guidance, while point trajectories supervise the attention mechanism to enforce fine-grained correspondence throughout denoising. We describe each component below.

\subsubsection{Event motion mask.}
\label{sec:int_mask}
We construct $\mathcal{M}_\text{event}$ at each frame timestamp $t$ to identify dynamic regions. We accumulate raw events over short symmetric windows $[t-\Delta, t+\delta]$ and $[t-\delta, t+\Delta]$, where $\Delta$ and $\delta$ define the temporal extents and $\delta \ll \Delta$. These windows capture events occurring shortly before and after the frame while maintaining a small overlap around $t$. From these, we compute past and future accumulation maps $\mathcal{A}_t^{-}(\mathbf{x})$ and $\mathcal{A}_t^{+}(\mathbf{x})$, and retain only pixels that are active in both:
\begin{equation}
\mathcal{M}_\text{event}(\mathbf{x}, t) = \mathbbm{1}\!\left[
\mathcal{A}_t^{-}(\mathbf{x}) > 0 \;\wedge\; \mathcal{A}_t^{+}(\mathbf{x}) > 0\right].
\end{equation}
The short temporal radius $\Delta$ biases the mask toward rapidly moving regions. Since this mask is derived entirely from raw events $\{e_i\}$ without learned components, it generalizes across sensor variants without adaptation. The mask is encoded through VAE and injected via channel concatenation into the denoising input, directing generation toward dynamic regions. The same construction is reused for evaluation in Sec.~\ref{sec:exp-oats}.

\subsubsection{Latent-space warping with optical flow.}
Given boundary frames $I_0$ and $I_1$, our estimator produces forward optical flow $\mathcal{F}_{0 \to t}$ from $I_0$ toward target timestamp $t$ and backward optical flow $\mathcal{F}_{1 \to t}$ from $I_1$ toward $t$. These boundary frames are encoded into latents $z_0$ and $z_1$, which are then warped to the target timestamp using the corresponding flow, producing warped latents $z_{0 \to t}$ and $z_{1 \to t}$. These are blended by temporal proximity, $z_{0 \to t}$ receiving higher weight near $I_0$, while $z_{1 \to t}$ dominates near $I_1$. This bidirectional warping and blending provides coarse spatial alignment as denoising initialization. This effectively reduces the generation problem from synthesizing content from scratch to refining a motion-aligned prior.

\subsubsection{Trajectory-guided attention.}
While the latent-level signals from optical flow warping and the event motion mask provide an initial motion-aligned condition, this coarse prior attenuates through successive denoising steps. To maintain strong geometric guidance throughout generation, we supervise the DiT's self-attention with bidirectional point trajectories $\mathcal{T}$ from TETO. At a selected attention layer and head, we sample query features $Q$ at trajectory positions in the boundary frame at timestamp $t_{\text{query}}$ and compute correlation against key features $K$ at all other frames to obtain predicted correspondence positions $\hat{\mathbf{x}}_t$ via soft argmax. We apply a Huber loss against the trajectory positions $\mathcal{T}_t$ only at visible frames ($v_t = 1$), excluding the query frame:
\begin{equation}
\mathcal{L}_{\text{attn}} = \frac{1}{|\mathcal{V}|}\sum_{t \in \mathcal{V}} 
\text{Huber}\!\left(\hat{\mathbf{x}}_t,\; \mathcal{T}_t\right),
\end{equation}
where $\mathcal{V} = \{t \mid v_t = 1,\; t \neq t_{\text{query}}\}$ . This loss is applied in both forward and backward directions and averaged across selected heads. More details on layer and head selection is provided in the Appendix.
\section{Experiments}
\label{sec:experiments}

We evaluate our motion estimator and VFI framework across three tasks on multiple real-world benchmarks. In line with our focus on real event statistics, we exclusively evaluate on real-world event datasets. We also introduce the Object-Adherent Trajectory Score (OATS) as a proxy metric to evaluate on highly dynamic dataset without ground-truth tracking annotation.

\subsubsection{Datasets.}
For point tracking, we benchmark on EVIMO2~\cite{burner2022evimo2}, which contains dynamic object motion with moving cameras. For optical flow, we evaluate on DSEC~\cite{gehrig2021dsec} in both zero-shot and in-domain settings, without ground-truth flow annotations, relying solely on pseudo-labels from our teacher for supervision. For video frame interpolation, we use BS-ERGB~\cite{tulyakov2022time} as the in-domain benchmark and HQ-EVFI~\cite{ma2024timelens} for cross-domain evaluation. BS-ERGB also serves as our primary evaluation set for tracking quality on highly dynamic scenes. We report OATS and additionally provide qualitative analysis to assess tracking fidelity. 

% imple detail. add to appendix
% We train TETO for 12 hours on a single NVIDIA RTX 5880 GPU with batch size 1. Each sample consists of 16 frames randomly cropped to 384 x 512. We use AdamW with learning rate $1e-5$. For video frame interpolation, we fine-tune a Wan2.1 DiT with LoRA on BS-ERGB. More detailed implementation details are provided in Appendix.
% VFI imple 추가하기 

\subsection{Event point tracking}
\begin{table}[t]
\centering
\begin{minipage}[t]{0.36\linewidth}
    \centering
    \captionof{table}{\textbf{Comparison on EVIMO2.}}
    \resizebox{\linewidth}{!}{
    \begin{tabular}{l|ccc}
    \toprule
    \multirow{2}{*}{\textbf{Methods}} & \multicolumn{3}{c}{\textbf{EVIMO2}}\\
     & \textbf{AJ↑} & \textbf{$\delta^{x}_{avg}$↑} & \textbf{OA↑} \\
    \midrule
    CoTracker~\cite{karaev2024cotracker} & 53.1 & 66.3 & 86.1  \\
    AllTracker~\cite{harley2025alltracker} & 57.9 & 72.4 & 91.7  \\
    \midrule
    ETAP~\cite{hamann2025etap} & \underline{66.1} & \underline{78.9} & \underline{89.5} \\
\hlrow{myblue}
    \Ours\ & \textbf{67.9} & \textbf{81.4} & \textbf{92.2} \\
    \bottomrule
    \end{tabular}}
    \label{table:track_evimo2_main}
\end{minipage}\hfill
\begin{minipage}[t]{0.60\linewidth}
    \vspace{+5pt}
    \centering
    \includegraphics[width=\linewidth]{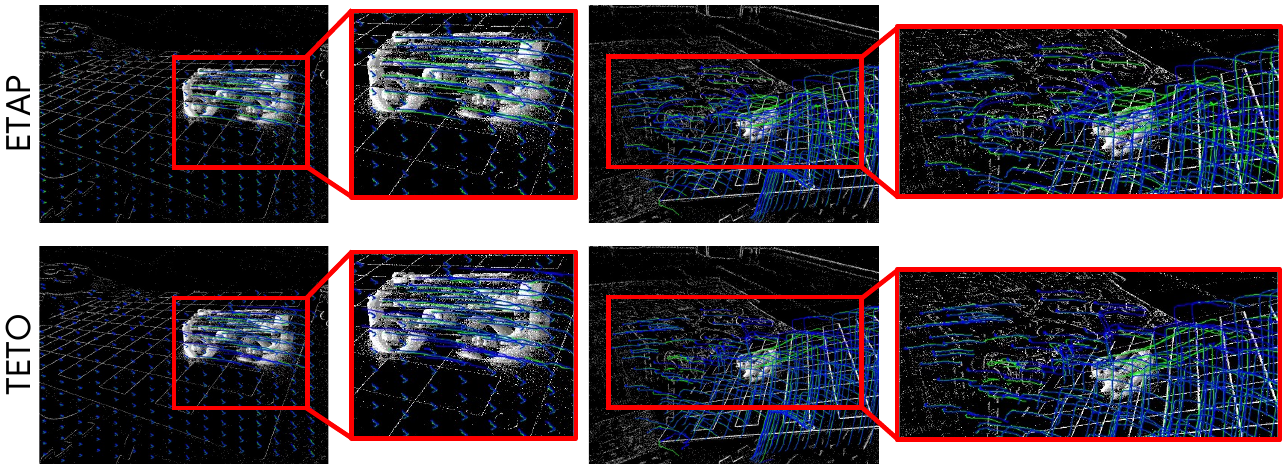}
    \captionof{figure}{\textbf{Qualitative Comparison.} Ground-truth trajectories are shown in green and model predictions in blue.}
    \vspace{-20pt}
    \label{fig:evimo_qual}
\end{minipage}
\end{table}
We evaluate on EVIMO2, which provides $640 \times 480$ event sequences featuring rigid objects with independent motion, including static scenes where event sparsity poses additional challenges. We use the extended EVIMO2 annotations from ETAP~\cite{hamann2025etap}, which provide longer trajectories and occlusion ground truth. To simulate an event-only setting, event streams are reconstructed into RGB frames using E2VID~\cite{rebecq2019events} for RGB tracker baselines~\cite{karaev2024cotracker, harley2025alltracker}.

As shown in Table~\ref{table:track_evimo2_main} and Figure~\ref{fig:evimo_qual}, our method achieves state-of-the-art results on all three TAP metrics, outperforming ETAP which trains on substantially larger synthetic data from EventKubric. Notably, our model is trained on only 25 minutes of real event video, yet it surpasses methods trained on hours of large-scale synthetic data.

\subsection{Event optical flow}
% requires: \usepackage{booktabs,multirow,array}
\newcolumntype{C}{w{c}{2.5em}}
\newlength{\seqcolwidth}
\settowidth{\seqcolwidth}{zurich\_city\_15\_a}
\newcolumntype{C}{w{c}{\dimexpr\seqcolwidth/2\relax}}
\begin{table*}[t]
\caption{\textbf{Quantitative comparison on DSEC.} Lower EPE, AE is better. \textbf{Bold} and \underline{underline} denote the best and second-best results within each category.}
\centering
\small
\resizebox{\textwidth}{!}{
\renewcommand{\arraystretch}{1.1}
% =========================
% Top block
% =========================
\begin{tabular}{l|c|CCCCCCCCCCCCCCCC}
\toprule
\multirow{2}{*}{Method} &\multirow{2}{*}{Setting} 
& \multicolumn{2}{c}{All}
& \multicolumn{2}{c}{interlaken\_00\_b}
& \multicolumn{2}{c}{interlaken\_01\_a}
& \multicolumn{2}{c}{thun\_01\_a}
& \multicolumn{2}{c}{thun\_01\_b}
& \multicolumn{2}{c}{zurich\_city\_12\_a}
& \multicolumn{2}{c}{zurich\_city\_14\_c}
& \multicolumn{2}{c}{zurich\_city\_15\_a} \\
\cmidrule(lr){3-4}\cmidrule(lr){5-6}\cmidrule(lr){7-8}\cmidrule(lr){9-10}\cmidrule(lr){11-12}\cmidrule(lr){13-14}\cmidrule(lr){15-16}\cmidrule(lr){17-18}
& & EPE$\downarrow$ & AE$\downarrow$ 
    & EPE$\downarrow$ & AE$\downarrow$ 
    & EPE$\downarrow$ & AE$\downarrow$ 
    & EPE$\downarrow$ & AE$\downarrow$ 
    & EPE$\downarrow$ & AE$\downarrow$ 
    & EPE$\downarrow$ & AE$\downarrow$ 
    & EPE$\downarrow$ & AE$\downarrow$ 
    & EPE$\downarrow$ & AE$\downarrow$  \\
\midrule
RTEF~\cite{brebion2021rtef} & In-domain
& 4.88 & -- 
& 8.59 & -- 
& 5.94 & -- 
& 3.01 & -- 
& 3.91 & --
& 3.14 & --
& 4.00 & --
& 3.78 & -- \\
MultiCM~\cite{shiba2022secrets} & In-domain
& 3.47 & 13.98 
& 5.74 & 9.19 
& 3.74 & 9.77 
& 2.12 & 11.06 
& 2.48 & 12.05
& 3.86 & 28.61
& 2.72 & 12.62
& 2.35 & 11.82\\
BTEB~\cite{paredes2021bteb} & In-domain
& 3.86 & --
& 6.32 & --
& 4.91 & --
& 2.33 & --
& 3.04 & --
& 2.62 & --
& 3.36 & --
& 2.97 & --\\
Paredes et al.~\cite{paredes2023taming} & In-domain
& 2.33 & 10.56
& 3.34 & 6.22
& 2.49 & 6.88
& 1.73 & 9.75
& 1.66 & 8.41 
& 2.72 & 23.16
& 2.64 & 10.23
& 1.69 & 8.88 \\
EV-FlowNet~\cite{zhu2019evflownet} & In-domain 
& 3.86 & -- 
& 6.32 & -- 
& 4.91 & -- 
& 2.33 & -- 
& 3.04 & --
& 2.62 & --
& 3.36 & --
& 2.97 & -- \\
MotionPriorCM~\cite{hamann2024motionpriorcm} & In-domain
& 3.20 & 8.53
& 3.21 & 4.89
& 2.38 & 5.46
& 1.39 & 6.99
& 1.54 & 6.55 
& 8.33 & 20.16
& 1.78 & 8.79
& 1.45 & 6.27\\
VSA-SM~\cite{you2024vsasm} & In-domain 
& 2.22 & 8.86
& 3.20 & 6.23
& 2.46 & 7.00
& 1.55 & 6.63
& 1.74 & 6.76 
& 2.19 & 17.13
& 1.69 & 7.57 
& 1.85 & 8.06 \\
E2FAI~\cite{guo2025unsupervised} & In-domain
& \underline{1.78} & 6.44
& \underline{3.08} & \underline{3.87}
& \underline{1.90} & \textbf{4.11}
& \underline{1.26} & 5.69
& \underline{1.15} & \underline{4.89}
& 1.92 & 14.35
& \underline{1.50} & 6.93
& \textbf{1.26} & \underline{5.46}\\
\cmidrule(lr){1-18}
\hlrow{myblue}
\textbf{TETO} & Zero-shot
& 2.15 & \underline{6.08}
& 4.08 & 6.27
& 2.19 & 5.45
& 1.56 & \underline{5.35}
& 1.78 & {5.36}
& \underline{1.35} & \underline{7.32}
& 1.63 & \underline{5.63}
& 2.00 & 6.57\\
\hlrow{myblue}
\textbf{TETO} & In-domain
& \textbf{1.39} & \textbf{4.31}
& \textbf{2.13} & \textbf{3.39}
& \textbf{1.51} & \underline{4.15}
& \textbf{1.04} & \textbf{3.90}
& \textbf{1.12} & \textbf{3.35}
& \textbf{1.06} & \textbf{6.12}
& \textbf{1.24} & \textbf{3.91}
& \underline{1.37} & \textbf{4.53}\\
\bottomrule
\end{tabular}
}
\vspace{10pt}
\label{table:of_main}
\end{table*}

We evaluate dense optical flow on DSEC without ground-truth flow annotations. We report two settings in Table~\ref{table:of_main}: zero-shot, where the estimator trained on 25 minutes recordings from EHPT-XC and without GT, is applied directly without any adaptation, and in-domain, where the estimator is further fine-tuned on DSEC using pseudo-labels from our teacher.
In the zero-shot setting, TETO already outperforms several dedicated optical flow methods that train on in-domain DSEC data, despite never seeing driving-domain data. Notably, our zero-shot angular error (6.08 AE) is already lower than E2FAI (6.44 AE), the previous best unsupervised method trained on in-domain data. With pseudo-label fine-tuning on DSEC, the estimator surpasses all compared unsupervised and self-supervised methods in both EPE and AE. These results demonstrate that our distillation-based motion estimator generalizes effectively across domains and that pseudo-label supervision, without any ground-truth flow, is sufficient to achieve state-of-the-art unsupervised optical flow.

\begin{figure*}[t!]
    \centering
    \includegraphics[width=1.0\linewidth]{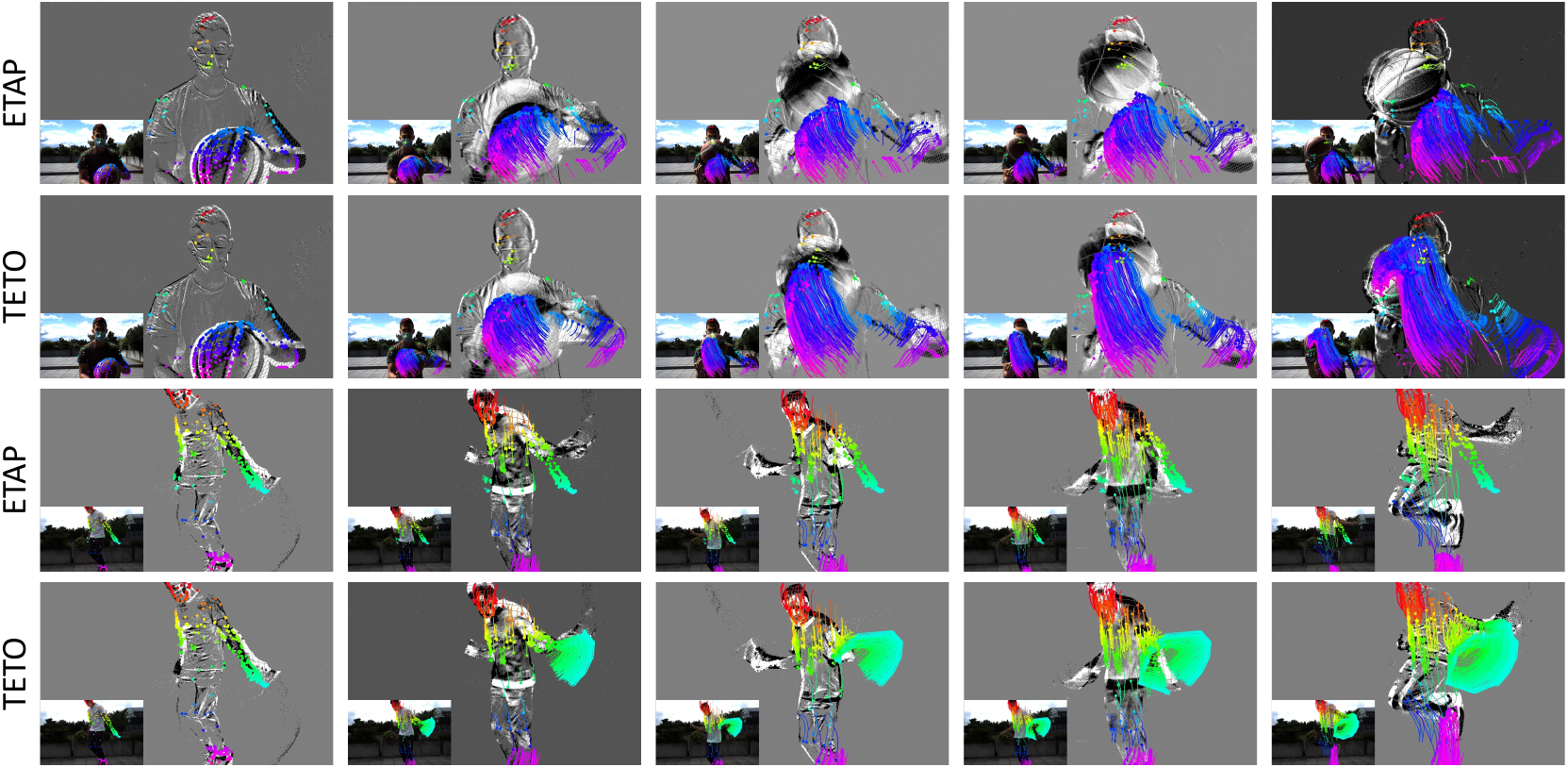}
    \caption{\textbf{Qualitative comparison under dynamic motion.}}
    % \vspace{-20pt}
    \label{fig:track_bsergb}
\end{figure*}
\begin{table}[t]
\centering
\caption{\textbf{OATS on BS-ERGB.}}
\vspace{-10pt}
\resizebox{0.8\linewidth}{!}
{
\begin{tabular}{l|ccccccc}
\toprule
\multirow{2}{*}{\textbf{Methods}} & \multicolumn{7}{c}{\textbf{BS-ERGB~\cite{tulyakov2022time}}}  \\
 & $\text{OATS}_{\text{0}}$↑ & $\text{OATS}_{\text{1}}$↑ & $\text{OATS}_{\text{2}}$↑ & $\text{OATS}_{\text{4}}$↑ & $\text{OATS}_{\text{8}}$↑ & $\text{OATS}_{\text{16}}$↑ & $\text{OATS}_{\text{avg}}$↑ \\
\midrule

ETAP~\cite{hamann2025etap}  & 0.7134 &  0.7256  & 0.7388  &  0.7568 &  0.7824  & 0.8149 &  0.7553  \\
\hlrow{myblue}
\Ours\  &    \textbf{0.8294} &  \textbf{0.8438} &  \textbf{0.8587} &  \textbf{0.8785} &  \textbf{0.9047} &  \textbf{0.9324} &  \textbf{0.8746}  \\
\bottomrule
\end{tabular}}
\label{table:track_bsergb_oats}
\end{table}

\subsection{Evaluation for highly dynamic scenes}
\label{sec:exp-oats}
\subsubsection{Tracking on dynamic scenes.}
Existing benchmarks contain limited motion dynamics, and reliable
annotation on highly dynamic scenes is infeasible. We therefore
evaluate on BS-ERGB through qualitative comparison
(Figure~\ref{fig:track_bsergb}) and a proposed proxy metric, the
Object-Adherent Trajectory Score (OATS,
Table~\ref{table:track_bsergb_oats}), which measures whether
tracked points remain on their associated objects over time using
SAM3~\cite{carion2025sam} segmentation.
$\mathcal{M}_\text{event}$ (Sec~\ref{sec:int_mask}) focuses
evaluation on dynamic regions. TETO significantly outperforms ETAP,
which struggles with complex real-world motion due to its synthetic
training. The detailed formulation of OATS is provided in the
Appendix.

\subsubsection{Extreme conditions.}
\begin{wrapfigure}{r}{0.4\linewidth}
    \centering
    \vspace{-25pt}
    \includegraphics[width=\linewidth]{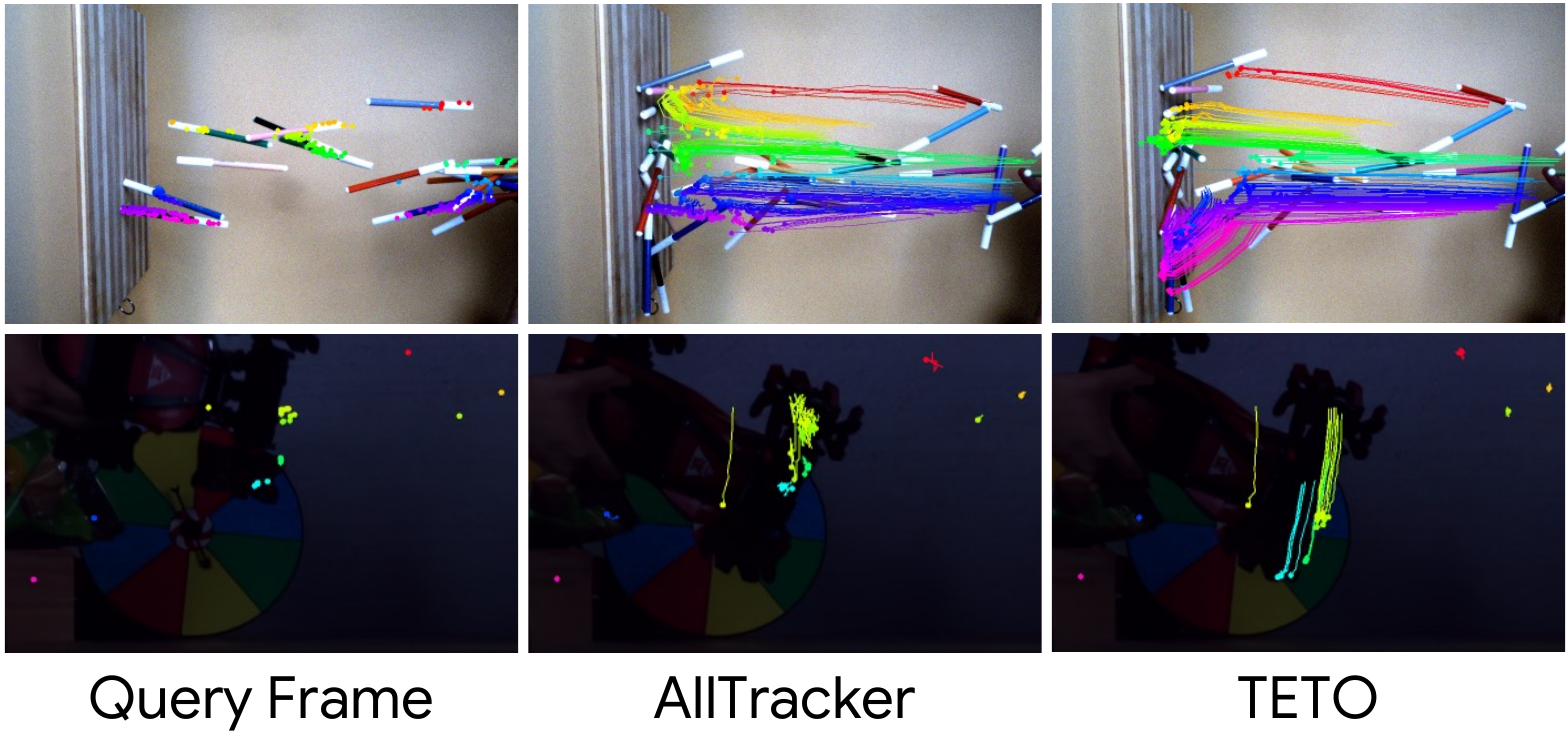}
    \vspace{-20pt}
    \caption{\textbf{Robust tracking in extreme conditions on BS-ERGB (Top) and EVFI-LL (Bottom)~\cite{zhang2024sim}.}}
    \vspace{-20pt}
    \label{fig:rgb_fail}
\end{wrapfigure}

Event cameras inherently excel where conventional sensors fail. As shown in Figure~\ref{fig:rgb_fail}, TETO produces coherent trajectories in nighttime and high-speed scenes where AllTracker, the teacher model, fails entirely. This is because our distillation objective (Sec~\ref{sec:tracking_tracker}) learns motion-aware representations from event temporal structure rather than replicating the teacher's RGB appearance, enabling the student to exploit temporal cues that only the event modality can provide.

\begin{figure*}[t!]
    \centering
    \includegraphics[width=1.0\linewidth]{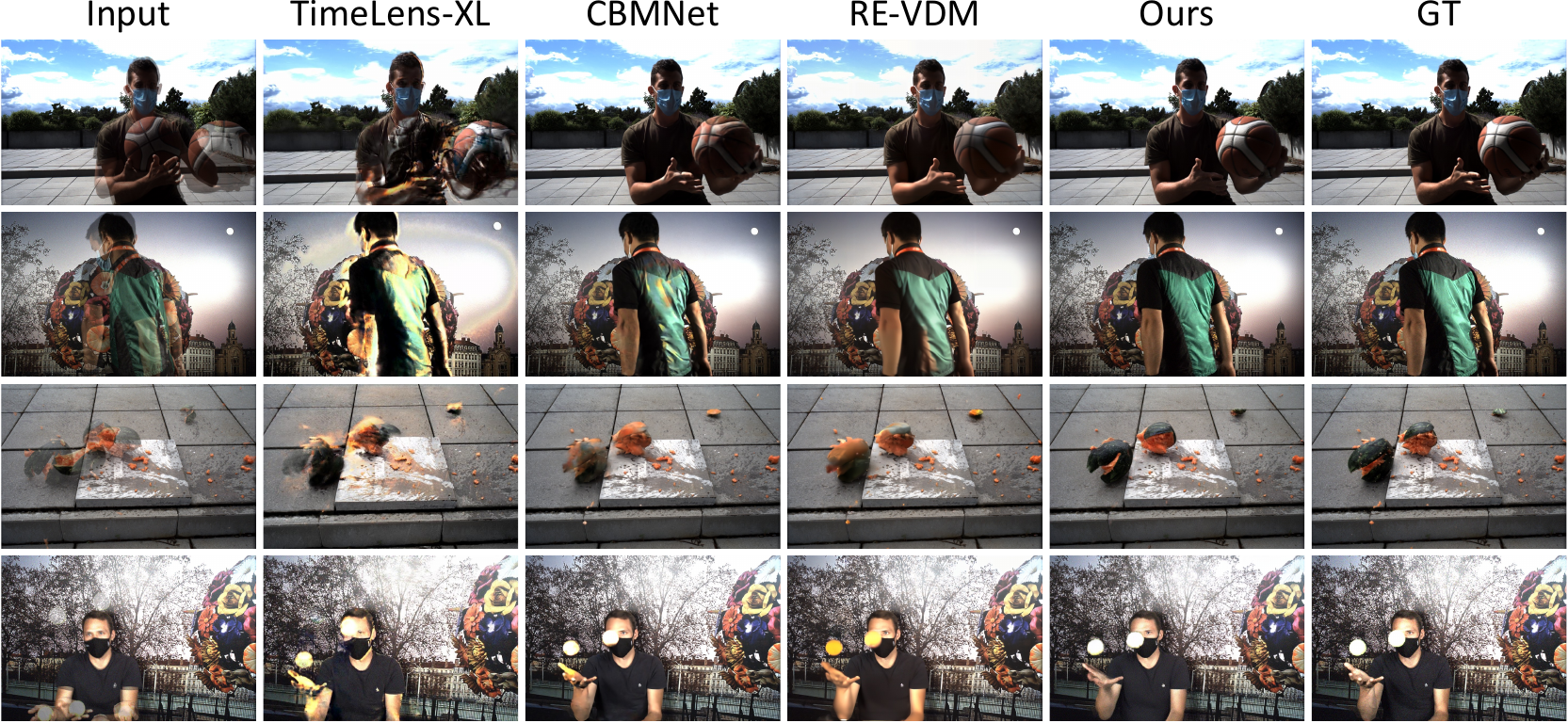}
    \caption{\textbf{Qualitative comparison under dynamic motion.}}\vspace{-10pt}
    \label{fig:vfi_bsergb}
\end{figure*}
\subsection{Video frame interpolation}

\begin{table*}[t]
\centering
\caption{\textbf{Video frame interpolation comparison on BS-ERGB}}
\label{tab:vfi_main}
\resizebox{\textwidth}{!}{%
\renewcommand{\arraystretch}{1.1}
\begin{tabular}{l|cccc|cccc|cccc}
\toprule
\multirow{2}{*}{\textbf{Method}}
& \multicolumn{4}{c|}{$\times$1}
& \multicolumn{4}{c|}{$\times$3}
& \multicolumn{4}{c}{$\times$6} \\
& FID$\downarrow$  & LPIPS$\downarrow$ & PSNR$\uparrow$ & SSIM$\uparrow$
& FID$\downarrow$  & LPIPS$\downarrow$ & PSNR$\uparrow$ & SSIM$\uparrow$
& FID$\downarrow$ & LPIPS$\downarrow$ & PSNR$\uparrow$ & SSIM$\uparrow$ \\
\midrule
TimeLens-XL~\cite{ma2024timelens}  
& 15.29 & \underline{0.0808} & \underline{28.2734} & 0.8223
& 38.11  & \underline{0.0695} & \textbf{29.7315} & \textbf{0.8367} 
&38.11 & 0.1314& 22.4263&0.7574 \\

CBMNet-Large~\cite{kim2023event} 
& \underline{10.64} & 0.1667 & \textbf{29.2348} & 0.7737 
& \underline{11.23} & 0.1730 & \underline{28.4550} &  0.7580
& \underline{13.31} & 0.1791 & \textbf{27.6728} & 0.7511 \\

RE-VDM~\cite{chen2025revdm} 
& 16.37 &0.1004 &28.0351 & \textbf{0.8631} 
& 16.66 &0.1268 & 27.0266& 0.8119
& 19.40 &\underline{0.1258} &\underline{26.9337} & \textbf{0.8426}\\

\midrule 
\hlrow{myblue}
\OursVFI
& \textbf{10.41} & \textbf{0.0684} & 25.7317 & \underline{0.8346}  
&\textbf{7.65}& \textbf{0.0689} & 26.2135 & \underline{0.8314}
& \textbf{7.88}& \textbf{0.0859 }&25.1521 &\underline{0.7955} \\
\bottomrule
\end{tabular}
}
\end{table*}

\begin{wraptable}{r}{0.6\linewidth}
\vspace{-35pt}

\centering
\caption{\textbf{Comparison on HQ-EVFI.}}
\label{table:vfi-cross}
\resizebox{\linewidth}{!}{%
\renewcommand{\arraystretch}{1.1}
\begin{tabular}{l|cccc|cccc}
\toprule

\multirow{2}{*}{\textbf{Method}}
& \multicolumn{4}{c|}{$\times$7}
& \multicolumn{4}{c}{$\times$15} \\
& FID$\downarrow$ & LPIPS$\downarrow$ & PSNR$\uparrow$ & SSIM$\uparrow$
& FID$\downarrow$ & LPIPS$\downarrow$ & PSNR$\uparrow$ & SSIM$\uparrow$ \\
\midrule
TimeLens-XL~\cite{ma2024timelens}  & 26.50 & 0.0601 & 25.5082 & 0.8856 & 28.83  & 0.0559 & 25.6656 & \textbf{0.8942}\\
CBMNet-Large~\cite{kim2023event} & 18.92 & 0.0717 & \textbf{29.0001} & 0.8849& 26.19 & 0.0827 & \textbf{27.3779} & 0.8767\\
RE-VDM~\cite{chen2025revdm} & 21.18  & 0.0695 & 26.9527 & 0.8714 & 33.42 & 0.0901 & 25.5576 & 0.8565 \\
\midrule
\hlrow{myblue}
\OursVFI & \textbf{17.39} & \textbf{0.0401} & 26.7525 & \textbf{0.9093} & \textbf{19.89} & \textbf{0.0501}  & 25.3575 & 0.8923 \\
\bottomrule
\end{tabular}
}
\vspace{-25pt}

\end{wraptable}

\subsubsection{Results on BS-ERGB.}

Table~\ref{tab:vfi_main} compares our method against both RGB-only and event-based VFI methods at $\times$1, $\times$3, and $\times$6 interpolation rates following evaluation protocol used in~\cite{chen2025revdm}. Our framework achieves the best perceptual quality across all rates in terms of FID and LPIPS. Qualitative results in Figure~\ref{fig:vfi_bsergb} demonstrate that our method produces sharper boundaries and more faithful reconstructions, particularly in dynamic motion regions where other methods exhibit blur or ghosting artifacts. The advantage is most pronounced in areas with independently moving objects, where precise motion conditioning from our estimator enables the DiT to synthesize appearance details rather than hallucinating displaced content.

\subsubsection{Zero-shot evaluation on HQ-EVFI.}

We also report frame interpolation generalization results in Table~\ref{table:vfi-cross}. The table presents the performance of models without fine-tuning on HQ-EVFI, highlighting their domain generalizability. Results are evaluated on a subset containing dynamic motions, which provides a more robust assessment. Performance on the full set, as well as the list of scene names included in the subset, is reported in the Appendix.

\subsection{Ablation study}
\begin{wraptable}{r}{0.4\linewidth}
\vspace{-35pt}
\caption{\textbf{Ablation study on EVIMO2.}}
\resizebox{\linewidth}{!}{
\begin{tabular}{lccc}
\toprule
\textbf{Methods} & \textbf{AJ↑} & \textbf{$\delta^{x}_{avg}$↑} & \textbf{OA↑} \\
\midrule
\Ours\ & \textbf{67.9} & \textbf{81.4} & \textbf{92.2} \\
\midrule
w/o $\mathcal{L}_{\text{flow}}$ & 66.3 & 79.9 & 92.1  \\
w/o temporal augmentation & 63.3 & 77.0 & 92.2  \\
w/o $\mathcal{M}_\mathrm{obj}$ & 60.3 & 74.6 & 91.5  \\
w/ feature loss & 63.7 & 76.6 & 91.9 \\
\bottomrule
\end{tabular}}
\vspace{-15pt}
\label{tab:track_evimo2_abl}
\end{wraptable}

\subsubsection{Point tracking.}
Table~\ref{tab:track_evimo2_abl} analyzes each component's contribution in TETO. The largest performance drop occurs when object-aware sampling is removed, causing the model to overfit to global ego-motion and fail to track independently moving objects, with trajectories collapsing to a single dominant motion. Removing flow supervision and disabling temporal augmentation each degrade performance, confirming that dense correspondence loss and multi-scale training both contribute to robust estimation. Introducing a feature loss that encourages RGB-like intermediate representations also degrades performance, supporting our claim in Sec~\ref{sec:tracking_tracker} that the distillation goal is to transfer matching mechanisms rather than replicate the teacher's visual encoding.

\begin{wraptable}{r}{0.4\linewidth}
\vspace{-30pt}
\caption{\textbf{Ablation study on BS-ERGB.}}
\resizebox{\linewidth}{!}{
\begin{tabular}{lcccc}
\toprule
\textbf{Methods}
& \textbf{FID$\downarrow$} 
& \textbf{LPIPS$\downarrow$} 
& \textbf{PSNR$\uparrow$} 
& \textbf{SSIM$\uparrow$} \\
\midrule
\multicolumn{5}{l}{\textit{Model Component}} \\
\OursVFI 
& \textbf{10.11} & \textbf{0.0821} & \textbf{25.4232} & \textbf{0.8022} \\
$-$ $\mathcal{M}_{\text{event}}$
& 11.06  & 0.0867 & 23.5839 & 0.8076 \\
$-$ $\mathcal{L}_{\text{attn}}$
& 10.82 & 0.0869 & 23.3661 & 0.8014\\
$-$ $\mathcal{F}_{\text{warp}}$
& 13.88  & 0.1203 & 21.9624  & 0.7573 \\
\midrule
\multicolumn{5}{l}{\textit{Motion Estimator}} \\
E2FAI~\cite{guo2025unsupervised}
& 13.34  & 0.0849 & 23.9826 & 0.7972 \\
\bottomrule
\end{tabular}}
\vspace{-20pt}
\label{tab:vfi_ablation}
\end{wraptable} 

\subsubsection{Video frame interpolation.}
Table~\ref{tab:vfi_ablation} analyzes each motion conditioning
component by progressive removal. Flow warping
($\mathcal{F}_{\text{warp}}$) provides coarse spatial alignment
and contributes the most. Attention supervision
($\mathcal{L}_{\text{attn}}$) refines correspondences beyond what
flow alone captures, improving perceptual quality. The event motion
mask ($\mathcal{M}_{\text{event}}$) directs generation capacity toward
dynamic regions. Each signal addresses a complementary aspect, and
replacing our estimator with E2FAI~\cite{guo2025unsupervised}
shows a clear performance gap, confirming that motion estimation
quality directly determines synthesis fidelity. All results are reported after training for 14,000 steps at $\times$3.
\section{Conclusion}
\label{sec:conclusion}
We presented TETO, a framework that learns event-based motion
estimation from only $\sim$25 minutes of real-world data through
RGB teacher distillation, achieving state-of-the-art tracking and
competitive optical flow without synthetic data. The resulting
motion estimates further enable high-quality video frame
interpolation, demonstrating that real data quality can effectively replace synthetic data scale for event-based motion estimation.

\clearpage
\appendix

\section*{Appendix Overview}

This appendix provides implementation details and additional results that complement the main paper.
Sec.~\ref{sec:supp_architecture} details the motion estimator architecture, including the motion-aware query sampling procedure, concentration network design, training configuration, and an ablation on the concentration network.
Sec.~\ref{sec:supp_vfi} describes the video frame interpolation architecture, covering the event motion mask construction, layer and head selection for attention supervision, and arbitrary-timestep generation.
Sec.~\ref{sec:supp_oats} formalizes the Object-Adherent Trajectory Score (OATS), including the evaluation protocol, scene selection criteria, and limitations.
Sec.~\ref{sec:supp_quantitative} presents additional quantitative results on InivTAP, DrivTAP, and the full HQ-EVFI benchmark, along with qualitative comparisons for point tracking and video frame interpolation.
Sec.~\ref{sec:supp_iei} provides the Inter-Event Interval (IEI) analysis comparing temporal statistics of real and synthetic events.
Sec.~\ref{sec:supp_limitations} discusses limitations of the current framework.

\renewcommand{\thesection}{\Alph{section}}
\renewcommand{\thefigure}{\arabic{figure}}
\renewcommand{\thetable}{\arabic{table}}

\setcounter{section}{0}
\setcounter{figure}{0}
\setcounter{table}{0}

\section{Motion estimator architecture}
\label{sec:supp_architecture}

\subsection{Details on motion-aware query sampling}
\label{sec:supp_motionmask}

We pre-compute a pool of motion-aware training samples before training begins, identifying crops that contain meaningful object motion and generating binary masks for query point allocation.

\subsubsection{Event-density crop selection.}
Since event activity is spatially non-uniform, we select crop positions based on event density rather than random cropping.
For each starting frame index, we divide the raw event frame into a $64{\times}64$ patch grid, count events per patch, and select the top-3 densest patches as candidate crop centers.
Each patch center is mapped to a $384{\times}512$ crop in RGB coordinates, accounting for the resolution difference between event and RGB sensors.

\subsubsection{Ego-motion decomposition.}
For each candidate crop, we run the frozen AllTracker teacher on paired RGB frames and decompose the predicted optical flow into camera ego-motion and object motion components.
We collect valid flow vectors across frames $t \in \{2, 4, 6, \ldots, 14\}$, filtering by visibility ($\geq 0.5$), confidence ($\geq 0.3$), and minimum flow magnitude ($\geq 0.5$ px).
When more than 20{,}000 valid points remain, we subsample with confidence-weighted probabilities.

We fit a global affine model $\mathbf{A}$ via two-pass RANSAC with a reprojection threshold of 2.0 px.
In the second pass, we discard the top 20\% of points by reprojection error and refit to obtain a more robust estimate.

\subsubsection{Object motion mask.}
Given the fitted affine model, we compute per-pixel residual flow magnitude at a reference frame:
\begin{equation}
r(\mathbf{x}) = \left\| \mathcal{F}_T(\mathbf{x}) - \mathbf{A}[\mathbf{x}, 1]^\top \right\|_2,
\end{equation}
weighted by a confidence gate $g(\mathbf{x}) = \mathbbm{1}[v(\mathbf{x}) \geq 0.5] \cdot \mathrm{clip}(c(\mathbf{x}), 0, 1)^{2.0}$, where $v$ and $c$ denote teacher visibility and confidence.
The gated residual $r_g = r \cdot g$ is thresholded at
\begin{equation}
\tau = \mathrm{median}(r_g) + 4.0 \times 1.4826 \times \mathrm{MAD}(r_g),
\end{equation}
where MAD is the median absolute deviation.
We apply morphological opening ($3{\times}3$ ellipse) and closing ($7{\times}7$ ellipse), then remove connected components smaller than 200 pixels.
A crop is retained in the motion-aware pool only if the mask covers at least 5\% of the crop area.

\subsubsection{Curated data pool.}
The strict filtering at each stage, from confidence-weighted flow selection through RANSAC-based decomposition to morphological cleanup, ensures that only samples with clearly identifiable object motion enter the training pool.
The curation produces a set of valid training entries, each containing the starting frame index, crop coordinates, and mask area ratio.
A single starting index may contribute up to 3 entries corresponding to different event-dense crop positions.
Sequences are annotated with their overall motion ratio (fraction of starting indices containing object motion), which is used as sampling weight during training.

\subsubsection{Query point allocation.}
During training, sequences are sampled with softmax-weighted probabilities (temperature $= 2.0$) based on their motion ratio, so that sequences with more frequent object motion are visited more often.
Given the pre-computed mask $\mathcal{M}_{\text{obj}}$ for the selected crop, we sample 90\% of query points from object motion regions and 10\% uniformly from the full frame.
Because the masks are derived from high-confidence teacher predictions with robust outlier rejection, the oversampled queries correspond to genuine object motion rather than noise or flow estimation artifacts.
This allocation prevents the model from overfitting to dominant ego-motion patterns while retaining background queries for global motion context.
Motion-aware sampling is activated from step 6{,}000 onward, allowing the model to first learn basic correspondences before focusing on dynamic regions.

\subsection{Motion estimator architecture}

\subsubsection{Concentration network.}
The concentration network $\mathbf{C}$ is a lightweight U-Net with base channels 16.
The encoder path consists of an initial block (10$\to$16 channels) followed by three downsampling stages (16$\to$32, 32$\to$64, 64$\to$128) using average pooling at stride 2.
The decoder path mirrors this structure with bilinear upsampling and skip connections, producing a 3-channel output through a final $1{\times}1$ convolution.
Each block contains three sequential Conv2d--BatchNorm--LeakyReLU(0.1) layers.
All weights are initialized with Kaiming normal and trained from scratch.
The concentration network adds approximately 0.6M parameters to the pipeline.

\subsubsection{Backbone encoder.}
We use the ConvNeXt-based encoder from AllTracker~\cite{harley2025alltracker}, which consists of three stages ([96$\to$192, 3 blocks], [192$\to$384, 3 blocks], [384, 9 blocks]) with a stride-4 stem convolution.
The encoder produces feature maps at $1/8$ spatial resolution.
All pretrained RGB weights are loaded, including the first convolutional layer, since the concentration network output already has 3 channels.

\subsubsection{Correlation and update block.}
We construct 5-level correlation volume pyramids with a search radius of 4, yielding 405-dimensional correlation features per query point.
The update block uses a GRU with hidden dimension 128 and runs $K{=}4$ refinement iterations.
Spatial propagation uses 2D convolutional blocks, and temporal reasoning is performed through pixel-aligned attention across a sliding window of 16 frames with stride 8.

\subsubsection{Event representation.}
At each timestamp, we collect the $N{=}300{,}000$ most recent events and construct $B{=}10$ temporal bins with exponentially increasing event counts following~\cite{nam2022stereo}.
Each bin accumulates event polarities at each pixel location to form a single-channel image.
We apply temporal stride augmentation during training, uniformly sampling strides from 1 to 4 for ERF-X170FPS (170 FPS) and from 1 to 2 for EHPT-XC ($\sim$20 FPS), forcing the model to handle varying motion magnitudes.

\subsubsection{Training.}
We train on a single NVIDIA RTX 5880 Ada GPU for 20{,}000 steps (approximately 12 hours) with batch size 1.
Each training sample consists of 16 frames randomly cropped to $384 {\times} 512$ from event-dense regions.
We use AdamW with a learning rate of $1{\times}10^{-5}$ and BF16 mixed precision.
All modules, including the concentration network, backbone encoder, correlation computation, and update block, are trained end-to-end.
The flow loss weight is $\lambda{=}0.01$ and the motion-aware sampling (Sec.~\ref{sec:supp_motionmask}) is activated from step 6{,}000 onward, allowing the model to first learn basic correspondences before focusing on dynamic regions.
The total number of trainable parameters is approximately 28M (0.6M for the concentration network and 27.4M for the AllTracker backbone).

\subsubsection{Pseudo-label generation.}
Pseudo-labels are generated online during training using a frozen AllTracker~\cite{harley2025alltracker} teacher operating on paired RGB frames with 4 refinement iterations.
The teacher jointly predicts trajectories, dense optical flow, per-point visibility, and confidence scores.
Rather than pre-filtering pseudo-labels with a hard threshold, we use the teacher's visibility and confidence outputs as soft weights in the training loss (Eq.~3 in the main paper), naturally down-weighting unreliable predictions.
Details on motion-aware data curation and query sampling are provided in Sec.~\ref{sec:supp_motionmask}.

\clearpage
\subsection{Ablation study}
\label{sec:supp_arch_ablation}

\begin{wraptable}{r}{0.5\linewidth}
\vspace{-30pt}
\caption{\textbf{Architecture design ablation on EVIMO2.}}
\resizebox{\linewidth}{!}{
\begin{tabular}{lccc}
\toprule
\textbf{Methods} & \textbf{AJ↑} & \textbf{$\delta^{x}_{avg}$↑} & \textbf{OA↑} \\
\midrule
\Ours\ & \textbf{67.9} & \textbf{81.4} & \textbf{92.2} \\
w/o ConcentNet \quad \quad & 56.5 & 69.4 & 91.6 \\
\bottomrule
\end{tabular}}
\vspace{-15pt}
\label{tab:arch_abl}
\end{wraptable}

A key design choice in TETO is the concentration network, $\mathcal{C}$, which converts $B$-channel event stacks into a 3-channel representation before the pretrained ConvNeXt encoder.
An alternative is to bypass the concentration network entirely and feed all $B{=}10$ event channels directly into the encoder by replacing its first convolutional layer with a zero-initialized 10-to-96 layer while keeping all other pretrained weights intact.
As shown in Table~\ref{tab:arch_abl}, removing the concentration network leads to a substantial performance drop.
With only $\sim$25 minutes of training data, the encoder cannot sufficiently learn its first layer from scratch, and the pretrained matching priors are lost due to the structural modification.
The concentration network, with only 0.6M additional parameters, avoids this by preserving the full pretrained encoder structure.
\section{Video frame interpolation architecture}
\label{sec:supp_vfi}

\subsubsection{Arbitrary-timestep generation.}
We set the VAE temporal compression ratio to 1, encoding each frame independently rather than jointly compressing multiple frames.
This allows our pipeline to generate an arbitrary number of intermediate frames at any target timestamp in a single forward pass, without being constrained by a fixed temporal compression window.

\subsubsection{Architecture and implementation details.}
\label{sec:impl}

We build upon the Wan2.1-Fun-V1.1-1.3B image-to-video diffusion model,
which employs a DiT backbone operating in VAE latent space at $\frac{1}{8}$
spatial resolution. The model is trained using a flow matching
objective with a rectified-flow scheduler. We fine-tune the DiT via
LoRA with rank $r{=}32$, while keeping the VAE
encoder/decoder, text encoder, and CLIP encoder frozen. When the attention
tracking loss is enabled, it is added to the diffusion loss with weight
$\lambda{=}0.01$.

We train on the BS-ERGB training split with a learning rate of
$10^{-4}$. Input frames are resized to $480 \times 832$. Interpolation ratios
are sampled from $\{1, 3, 6, 14\}$.

\subsubsection{Layer and head selection.}

\begin{figure}
    \centering
    \includegraphics[width=0.8\linewidth]{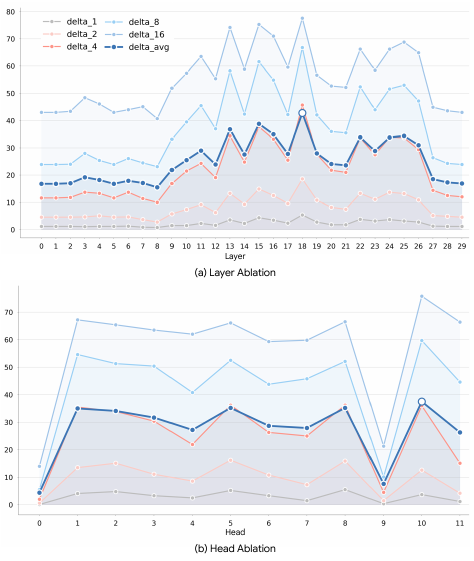}
    \caption{\textbf{Layer and head ablation for matching property.}}
    \label{fig:suppl_layer_ablation}
\end{figure}

Since our attention supervision operates directly on the transformer attention maps, it is applicable only to models with a DiT-style attention structure. We follow ~\cite{nam2025emergent, son2025repurposing} to select the layer and attention heads used for attention supervision. First, we perform a layer ablation using zero-shot matching on a subset of DAVIS~\cite{doersch2022tap}. Based on this analysis, we select the best-performing layer (Layer 18). We then conduct a head ablation within this layer and choose the top-$k$ heads ($k=3$) with the best matching performance. Figure~\ref{fig:suppl_layer_ablation}(a) shows the layer-wise matching performance, and (b) shows the head-wise results within the selected layer. Based on these results, we select heads 5, 8, and 10 from Layer 18.

\subsubsection{Event motion mask construction.}
\label{sec:supp_event_mask}
\begin{figure}
    \centering
    \includegraphics[width=\linewidth]{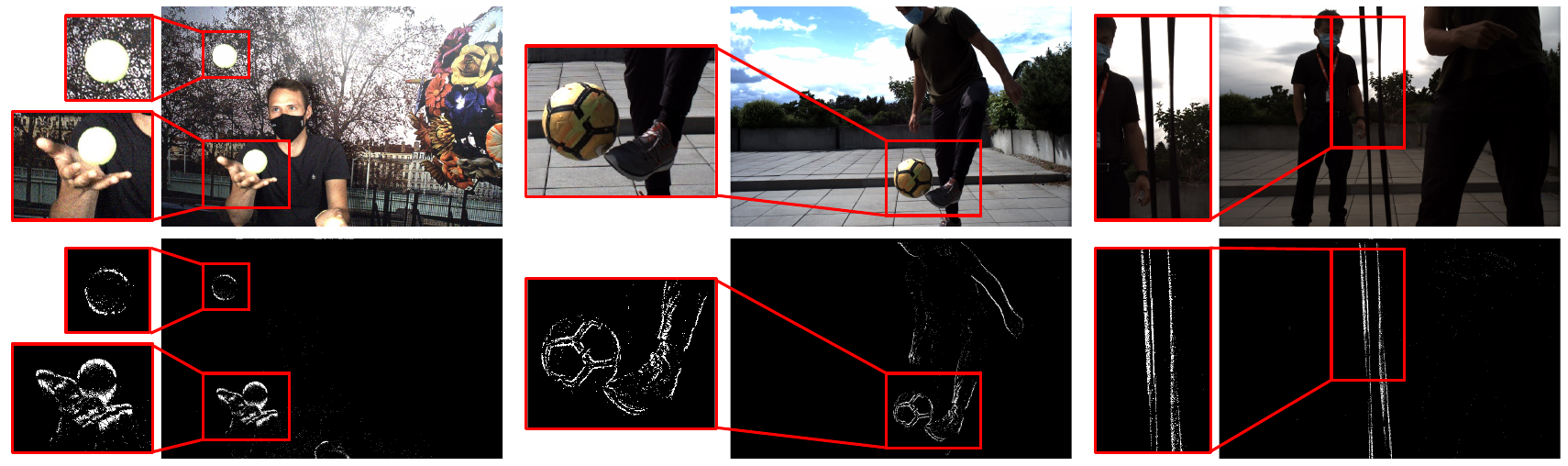}
    \caption{\textbf{Visualization of Event Motion Mask $\mathcal{M}_{\text{event}}$.}}
\label{fig:supp_motionmask_qual}
\end{figure}

The event motion mask $\mathcal{M}_{\text{event}}$ (Eq.~6 in the main paper)
identifies dynamic regions at each frame timestamp using only raw events,
without any learned components or RGB input.
While Eq.~6 presents the simplified single-scale intersection form,
the full construction operates over two temporal scales to
balance sensitivity and noise suppression.
We construct the mask by accumulating events over two temporal scales and combining them to isolate fast-moving regions.

For each frame at timestamp $t$, we collect two pairs of event stacks from symmetric temporal windows around $t$.
The \emph{wide window} accumulates the $N_{\text{wide}}\\{=}10{,}000$ most recent events before and after $t$, capturing broadly active regions.
The \emph{narrow window} accumulates the $N_{\text{narrow}}{=}1{,}000$ events closest to $t$, capturing only regions with rapid motion near the frame boundary.
Each stack is converted into a binary activation map indicating whether any event fired at each pixel.

The final mask combines these scales as follows.
Narrow-window activations from the current and previous frames are unioned to capture pixels that fired close to either frame boundary, ensuring fast-moving objects are detected even if they appear in only one direction.
Wide-window activations from both frames are intersected to retain only persistently moving regions, filtering out transient noise.
The sharp mask is the union of these two terms:
\begin{equation}
\mathcal{M}_{\text{event}} = (\mathcal{A}_{\text{narrow}}^{-} \cup \mathcal{A}_{\text{narrow}}^{+}) \cup (\mathcal{A}_{\text{wide}}^{-} \cap \mathcal{A}_{\text{wide}}^{+}),
\end{equation}
where $\mathcal{A}^{-}$ and $\mathcal{A}^{+}$ denote activations from the previous and current frame windows, respectively.

This two-scale design ensures that the mask captures fast object motion sharply while suppressing static background and sensor noise.
As shown in Figure~\ref{fig:supp_motionmask_qual}, the resulting mask tightly covers independently moving objects without relying on any motion estimation or segmentation model.
The same mask construction is used both for directing VFI generation capacity toward dynamic regions and for focusing OATS evaluation on moving objects.
\section{Object-Adherent Trajectory Score (OATS)}
\label{sec:supp_oats}

\begin{figure}
    \centering
    \includegraphics[width=0.6\linewidth]{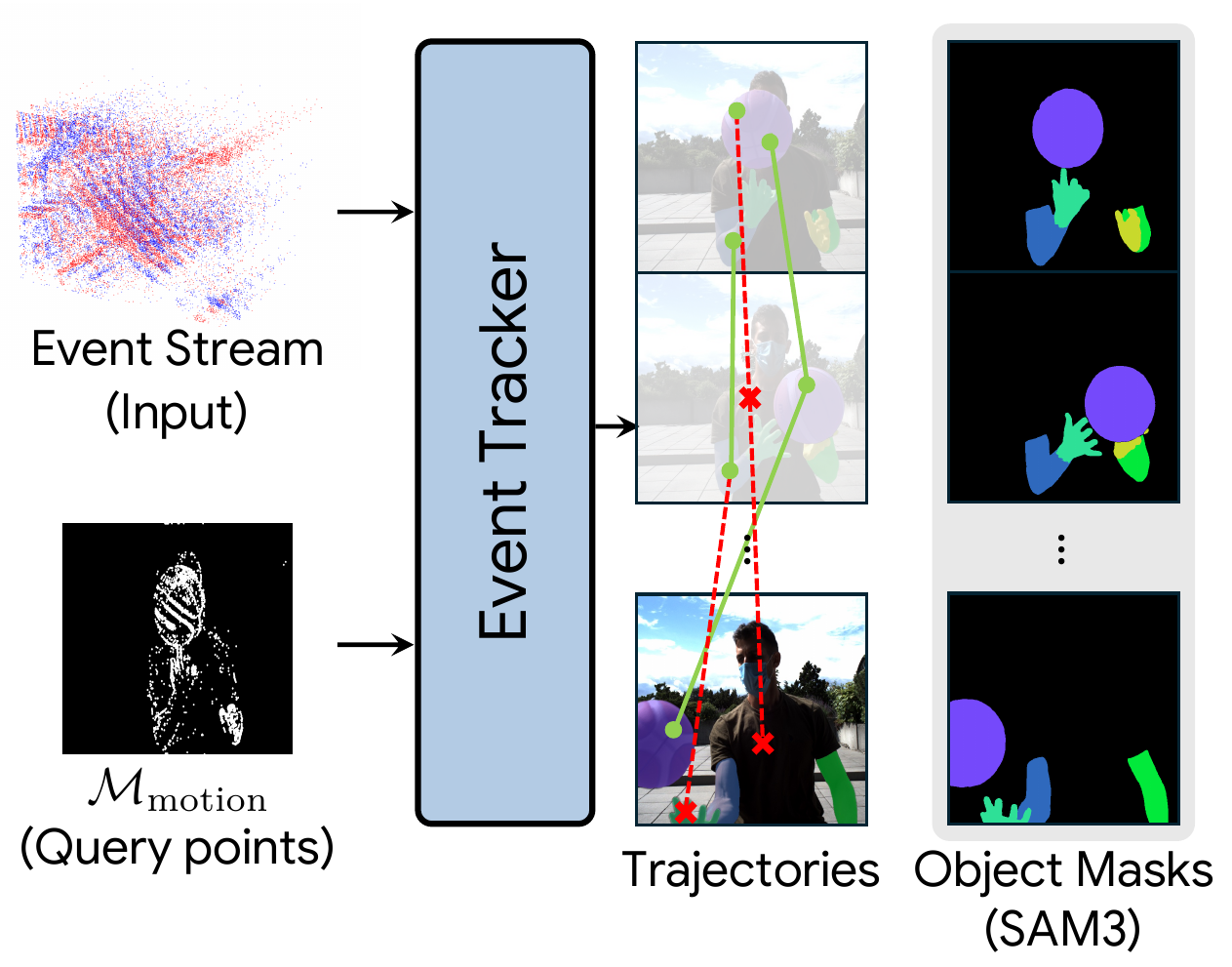}
    \caption{\textbf{OATS evaluation pipeline.} Motion regions are extracted solely from events, and then query trajectories are evaluated based on whether they remain within the corresponding object masks over time.}
    \label{fig:suppl_oats}
    \vspace{-10pt}
\end{figure}

\subsubsection{Object-Adherent Trajectory Score.}
To evaluate tracking quality without ground-truth trajectories, we leverage object segmentation as a proxy. We obtain per-frame object masks $\{{O}_i^t\}$ using SAM3~\cite{carion2025sam} where $i$ indexes objects and $t$ indexes frames. Query points are sampled from the intersection of $\mathcal{M}_{\text{event}}$ and the object masks at the first frame. With slight abuse of notation, we denote $O_j^t$ as the mask of the object from which query $j$ originates at frame $t$.

OATS measures how consistently a tracked point remains on its associated object. Following standard protocols, we evaluate under multiple distance thresholds $\delta \in \Delta$, where $\Delta = \{0, 1, 2, 4, 8, 16\}$:
\begin{equation}
\text{OATS}_\delta = \frac{1}{|Q|} \sum_{j=1}^{|Q|} \frac{1}{|\mathcal{V}_j|} \sum_{t \in \mathcal{V}_j}
\mathbbm{1} \Big[ d\big(\mathcal{T}_j^t, O_j^t\big) \le \delta \Big],
\end{equation}
where $\mathcal{T}_j^t$ denotes the predicted location of query $j$ at frame $t$, $\mathcal{V}_j$ is the set of visible frames for query $j$, and $d(\cdot, O)$ is the Euclidean distance to the nearest pixel in mask $O$. We report $\text{OATS}_{\text{avg}}$ as the mean across all thresholds.

\subsubsection{Per-scene OATS.}

\begin{table*}[htp!]
\centering
\caption{\textbf{Per-scene OATS comparison on BS-ERGB dataset.}}
\resizebox{0.87\linewidth}{!}{
\begin{tabular}{l|c|ccccccc}
\toprule
\multirow{2}{*}{\textbf{Scene Name}} & \multirow{2}{*}{\textbf{Model}} & \multicolumn{7}{c}{\textbf{BS-ERGB}} \\
 &  & $\text{OATS}_{\text{0}}$↑ & $\text{OATS}_{\text{1}}$↑ & $\text{OATS}_{\text{2}}$↑ & $\text{OATS}_{\text{4}}$↑ & $\text{OATS}_{\text{8}}$↑ & $\text{OATS}_{\text{16}}$↑ & $\text{OATS}_{\text{avg}}$↑ \\
 \midrule

\multirow{2}{*}{\texttt{ball\_05}} & ETAP & 0.4538 & 0.4677 & 0.4840 & 0.5095 & 0.5518 & 0.6234 & 0.5150 \\
 & \Ours\ & \textbf{0.7131} & \textbf{0.7321} & \textbf{0.7542} & \textbf{0.7872} & \textbf{0.8357} & \textbf{0.8976} & \textbf{0.7867} \\
\midrule
\multirow{2}{*}{\texttt{ball\_06}} & ETAP & 0.6553 & 0.6684 & 0.6831 & 0.7076 & 0.7456 & 0.8085 & 0.7114 \\
 & \Ours\ & \textbf{0.7602} & \textbf{0.7747} & \textbf{0.7927} & \textbf{0.8209} & \textbf{0.8633} & \textbf{0.9168} & \textbf{0.8214} \\
\midrule
\multirow{2}{*}{\texttt{basket\_08}} & ETAP & 0.6785 & 0.6851 & 0.6930 & 0.7068 & 0.7314 & 0.7694 & 0.7107 \\
 & \Ours\ & \textbf{0.7926} & \textbf{0.8015} & \textbf{0.8121} & \textbf{0.8285} & \textbf{0.8561} & \textbf{0.8945} & \textbf{0.8309} \\
\midrule
\multirow{2}{*}{\texttt{basket\_09}} & ETAP & 0.6781 & 0.6829 & 0.6892 & 0.6993 & 0.7166 & 0.7467 & 0.7021 \\
 & \Ours\ & \textbf{0.8135} & \textbf{0.8219} & \textbf{0.8327} & \textbf{0.8490} & \textbf{0.8768} & \textbf{0.9073} & \textbf{0.8502} \\
\midrule
\multirow{2}{*}{\texttt{eggs\_04}} & ETAP & 0.9539 & 0.9584 & 0.9621 & 0.9656 & 0.9692 & 0.9738 & 0.9638 \\
 & \Ours\ & \textbf{0.9794} & \textbf{0.9842} & \textbf{0.9875} & \textbf{0.9910} & \textbf{0.9937} & \textbf{0.9965} & \textbf{0.9887} \\
\midrule
\multirow{2}{*}{\texttt{football\_04}} & ETAP & 0.6541 & 0.6670 & 0.6831 & 0.7065 & 0.7478 & 0.8061 & 0.7108 \\
 & \Ours\ & \textbf{0.7634} & \textbf{0.7796} & \textbf{0.7981} & \textbf{0.8258} & \textbf{0.8654} & \textbf{0.9174} & \textbf{0.8250} \\
\midrule
\multirow{2}{*}{\texttt{horse\_11}} & ETAP & \textbf{0.7777} & \textbf{0.8035} & \textbf{0.8306} & \textbf{0.8631} & \textbf{0.9022} & \textbf{0.9445} & \textbf{0.8536} \\
 & \Ours\ & 0.7476 & 0.7747 & 0.8019 & 0.8349 & 0.8759 & 0.9134 & 0.8247 \\
\midrule
\multirow{2}{*}{\texttt{horse\_12}} & ETAP & 0.8729 & 0.8929 & 0.9152 & 0.9424 & 0.9699 & 0.9891 & 0.9304 \\
 & \Ours\ & \textbf{0.8951} & \textbf{0.9152} & \textbf{0.9369} & \textbf{0.9606} & \textbf{0.9823} & \textbf{0.9959} & \textbf{0.9477} \\
\midrule
\multirow{2}{*}{\texttt{horse\_13}} & ETAP & 0.8607 & 0.8787 & 0.9009 & 0.9308 & 0.9658 & 0.9898 & 0.9211 \\
 & \Ours\ & \textbf{0.8796} & \textbf{0.8988} & \textbf{0.9208} & \textbf{0.9498} & \textbf{0.9797} & \textbf{0.9958} & \textbf{0.9374} \\
\midrule
\multirow{2}{*}{\texttt{horse\_18}} & ETAP & 0.9481 & 0.9625 & 0.9732 & 0.9835 & \textbf{0.9938} & \textbf{0.9990} & 0.9767 \\
 & \Ours\ & \textbf{0.9506} & \textbf{0.9652} & \textbf{0.9769} & \textbf{0.9871} & 0.9935 & 0.9989 & \textbf{0.9787} \\
\midrule
\multirow{2}{*}{\texttt{horse\_20}} & ETAP & 0.8392 & 0.8498 & 0.8615 & 0.8777 & 0.9050 & 0.9350 & 0.8780 \\
 & \Ours\ & \textbf{0.8562} & \textbf{0.8674} & \textbf{0.8801} & \textbf{0.8982} & \textbf{0.9220} & \textbf{0.9426} & \textbf{0.8944} \\
\midrule
\multirow{2}{*}{\texttt{jacket\_03}} & ETAP & 0.7582 & 0.7732 & 0.7913 & 0.8199 & 0.8623 & 0.9105 & 0.8192 \\
 & \Ours\ & \textbf{0.8584} & \textbf{0.8724} & \textbf{0.8851} & \textbf{0.9036} & \textbf{0.9322} & \textbf{0.9642} & \textbf{0.9027} \\
\midrule
\multirow{2}{*}{\texttt{juggling\_06}} & ETAP & 0.3120 & 0.3175 & 0.3238 & 0.3340 & 0.3503 & 0.3757 & 0.3356 \\
 & \Ours\ & \textbf{0.6921} & \textbf{0.7014} & \textbf{0.7141} & \textbf{0.7331} & \textbf{0.7645} & \textbf{0.7973} & \textbf{0.7338} \\
\midrule
\multirow{2}{*}{\texttt{may28\_axe\_01}} & ETAP & \textbf{0.9489} & \textbf{0.9596} & \textbf{0.9692} & \textbf{0.9759} & \textbf{0.9815} & \textbf{0.9840} & \textbf{0.9698} \\
 & \Ours\ & 0.9475 & 0.9591 & 0.9662 & 0.9711 & 0.9739 & 0.9753 & 0.9655 \\
\midrule
\multirow{2}{*}{\texttt{may29\_redbull\_01}} & ETAP & 0.6605 & 0.6757 & 0.6909 & 0.7097 & 0.7411 & 0.7876 & 0.7109 \\
 & \Ours\ & \textbf{0.7671} & \textbf{0.7871} & \textbf{0.8073} & \textbf{0.8328} & \textbf{0.8691} & \textbf{0.9043} & \textbf{0.8280} \\
\midrule
\multirow{2}{*}{\texttt{rope\_jumping\_01}} & ETAP & 0.5744 & 0.5900 & 0.6078 & 0.6335 & 0.6776 & 0.7389 & 0.6370 \\
 & \Ours\ & \textbf{0.7653} & \textbf{0.7837} & \textbf{0.8038} & \textbf{0.8340} & \textbf{0.8761} & \textbf{0.9272} & \textbf{0.8317} \\
\midrule
\multirow{2}{*}{\texttt{street\_crossing\_08}} & ETAP & \textbf{0.9329} & \textbf{0.9437} & \textbf{0.9534} & \textbf{0.9630} & \textbf{0.9699} & \textbf{0.9739} & \textbf{0.9561} \\
 & \Ours\ & 0.8980 & 0.9094 & 0.9207 & 0.9332 & 0.9455 & 0.9562 & 0.9272 \\
\midrule
\multirow{2}{*}{\texttt{watermelon\_01}} & ETAP & 0.2820 & 0.2835 & 0.2864 & 0.2928 & 0.3007 & 0.3116 & 0.2928 \\
 & \Ours\ & \textbf{0.8498} & \textbf{0.8594} & \textbf{0.8658} & \textbf{0.8722} & \textbf{0.8786} & \textbf{0.8818} & \textbf{0.8679} \\
\bottomrule
\end{tabular}
}
\label{table:suppl_oats}
\end{table*}

Table~\ref{table:suppl_oats} reports OATS for each evaluated scene in BS-ERGB. For each scene, we show results across multiple distance thresholds $\delta$, as well as the average score $\text{OATS}_{\text{avg}}$.

\subsubsection{Scene selection.}

Not all sequences in BS-ERGB are suitable for reliable object-level evaluation using segmentation masks. We therefore filter scenes according to the following criteria: (1) fluid-dominated scenes, where most of the motion arises from liquids or flames, are excluded because such regions lack well-defined object boundaries and produce unstable segmentation masks; (2) scenes with severe motion blur in RGB frames are removed since blur degrades segmentation quality and leads to unreliable object masks; and (3) sequences containing many visually similar small objects are excluded when SAM3 fails to consistently distinguish instance identities. The final evaluation set consists only of scenes where object masks are visually verified to be spatially and temporally consistent.

\subsubsection{Segmentation mask generation and refinement.}

\begin{figure}[t]
    \centering
    \includegraphics[width=0.8\linewidth]{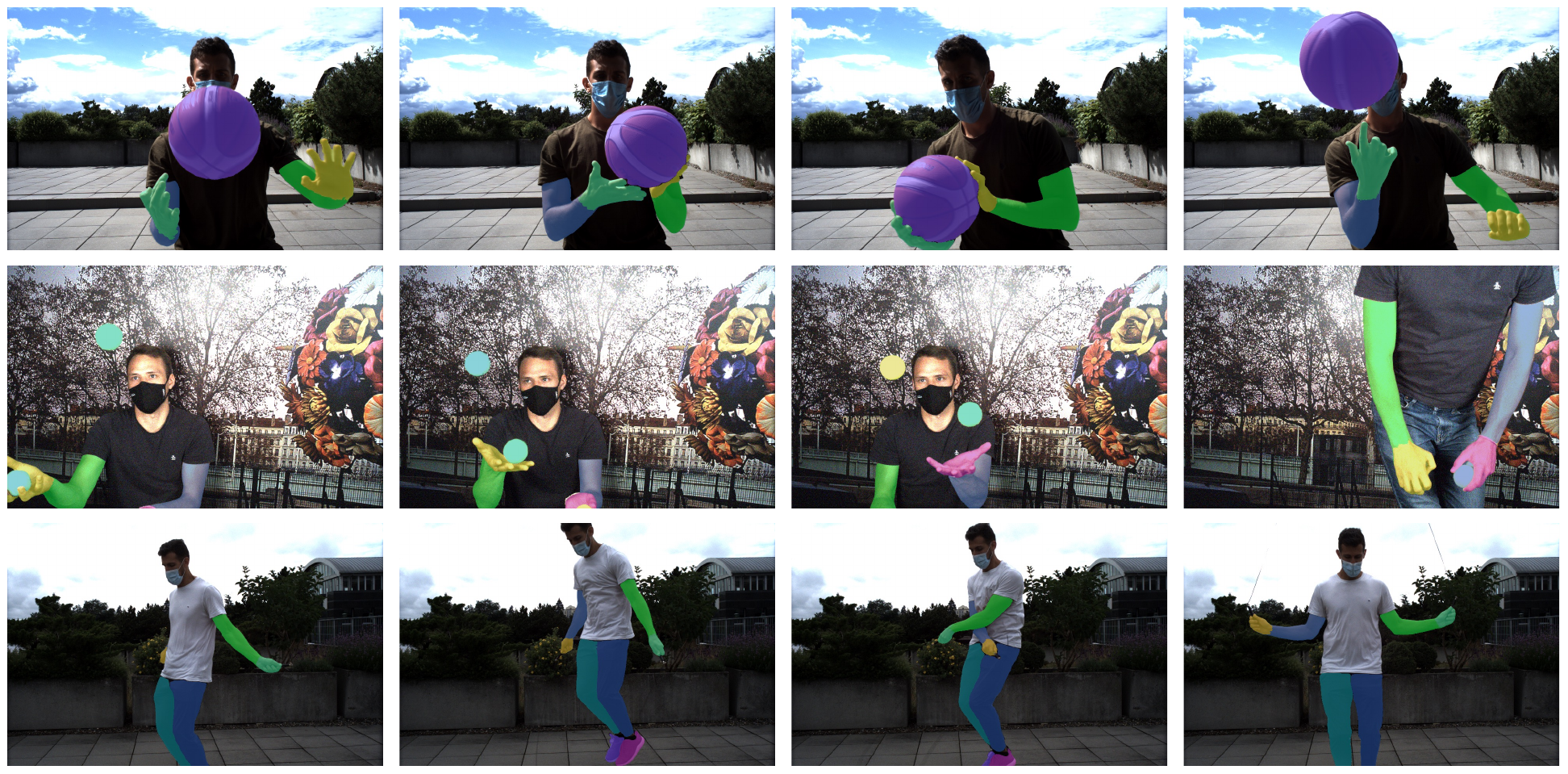}
    \caption{\textbf{Visualization of SAM3 video segmentation masks.}}
    \label{fig:suppl_sam_mask}
\vspace{-15pt}
\end{figure}
Per-frame object segmentation masks are generated from the RGB frames using SAM3 with manual text prompts. Following automatic mask generation, we conduct a manual verification and refinement process: masks that are incorrect or missing are removed, and for objects that are correctly segmented but assigned inconsistent mask IDs across frames, we manually correct the IDs to ensure temporal consistency. This procedure guarantees that each tracked trajectory is consistently associated with its corresponding object throughout the sequence. 
% Examples of the resulting masks are shown in Figure~\ref{fig:suppl_sam_mask}.

\subsubsection{Distance threshold implementation.}

For each distance threshold $\delta$, we determine whether a predicted point lies sufficiently close to its associated object mask. In practice, instead of computing explicit point-to-mask distances, we dilate the binary object mask by $\delta$ pixels and check whether the predicted point falls inside the dilated region.

Formally, for an object mask $O_j^t$ at time $t$, we construct a dilated mask $O_j^t(\delta)$ using morphological dilation with a circular structuring element of radius $\delta$. A prediction $\mathcal{T}_j^t$ is considered correct if it lies inside $O_j^t(\delta)$.

This implementation is equivalent to checking whether the Euclidean distance between the point and its nearest mask pixel is less than or equal to $\delta$.

\subsubsection{Limitation of OATS.}
Although OATS provides a meaningful proxy for evaluating how consistently tracked points adhere to their originating objects in unannotated or highly challenging sequences, where ground-truth trajectories are unavailable and even state-of-the-art RGB trackers fail, it has inherent limitations. In particular, OATS is computed only over frames where predicted points are marked as visible and does not directly evaluate the accuracy of visibility prediction. 

\section{Quantitative and qualitative results}
\label{sec:supp_quantitative}

\begin{table}
\vspace{-15pt}
\centering
\caption{\textbf{Quantitative results on real-world TAP benchmark.}}
\label{table:track-tapformer}
\resizebox{1.0\linewidth}{!}{%
\renewcommand{\arraystretch}{1.1}
\begin{tabular}{l|cc|ccccccc}
\toprule

\multirow{2}{*}{\textbf{Method}}
& \multirow{2}{*}{\makecell{Training\\data}} & \multirow{2}{*}{\makecell{Total\\recording}}
& \multicolumn{3}{c}{InivTAP} & \multicolumn{3}{c}{DrivTAP} & \multirow{2}{*}{\textbf{Mean$\delta^{x}_{avg}$↑}}\\
&& & \textbf{AJ↑} & \textbf{$\delta^{x}_{avg}$↑} & \textbf{OA↑}
& \textbf{AJ↑} & \textbf{$\delta^{x}_{avg}$↑} & \textbf{OA↑}\\
\midrule
E2Vid-PIPs++~\cite{zheng2023pointodyssey} & - & - &- &14.3 &- &- &30.4 &- &22.4 \\
E2Vid-Chrono~\cite{kim2025exploring} & - & - &10.4 &17.9 &68.8 &8.1 &16.9 &68.7 &17.4\\
E2Vid-CoTracker3~\cite{karaev2025cotracker3} & - & - &12.8 &19.4 &56.6 &18.6 &27.0 &\underline{89.4} &23.2 \\
ETAP~\cite{hamann2025etap}   & synthetic & 5 h &12.8 &22.3 &\textbf{86.3} &13.5 &27.8 &68.1 &25.1 \\
TAPFormer-E~\cite{liu2026tapformer}  & synthetic & 6 h &\textbf{20.9} &\textbf{29.8} &79.0 &28.9 &37.6 &\textbf{90.2} &33.7 \\
\midrule
\hlrow{myblue}
\Ours  & real & 25 min &\underline{17.3} &\underline{29.2} &\underline{83.0} &\textbf{33.1} &\textbf{45.5} &87.2 &\textbf{37.4} \\
% \Ours  &\underline{18.3} &\textbf{30.5} &\underline{83.5} &\textbf{33.1} &\textbf{45.5} &87.2 &\textbf{38.0} \\
\bottomrule
\end{tabular}
}
\vspace{-15pt}
\end{table}

\subsection{Results on InivTAP and DrivTAP}
\label{sec:supp_inivtap}
We additionally evaluate on InivTAP and DrivTAP, two real-world benchmarks recently introduced alongside TAPFormer-E~\cite{liu2026tapformer}, both published after our main paper submission.
InivTAP contains indoor sequences captured with a DAVIS event camera at 346$\times$260 resolution, while DrivTAP consists of outdoor driving sequences with diverse traffic dynamics.

As shown in Table~\ref{table:track-tapformer}, TETO achieves the best overall mean $\delta^x_{\text{avg}}$ across both benchmarks.
On DrivTAP, TETO outperforms all compared methods including TAPFormer-E.
On InivTAP, TETO achieves comparable $\delta^x_{\text{avg}}$ to TAPFormer-E while showing a gap in AJ, which we attribute to the resolution difference between our training data (1280$\times$720) and InivTAP (346$\times$260).

\begin{figure}[t!]
    \centering
    \includegraphics[width=.8\linewidth]{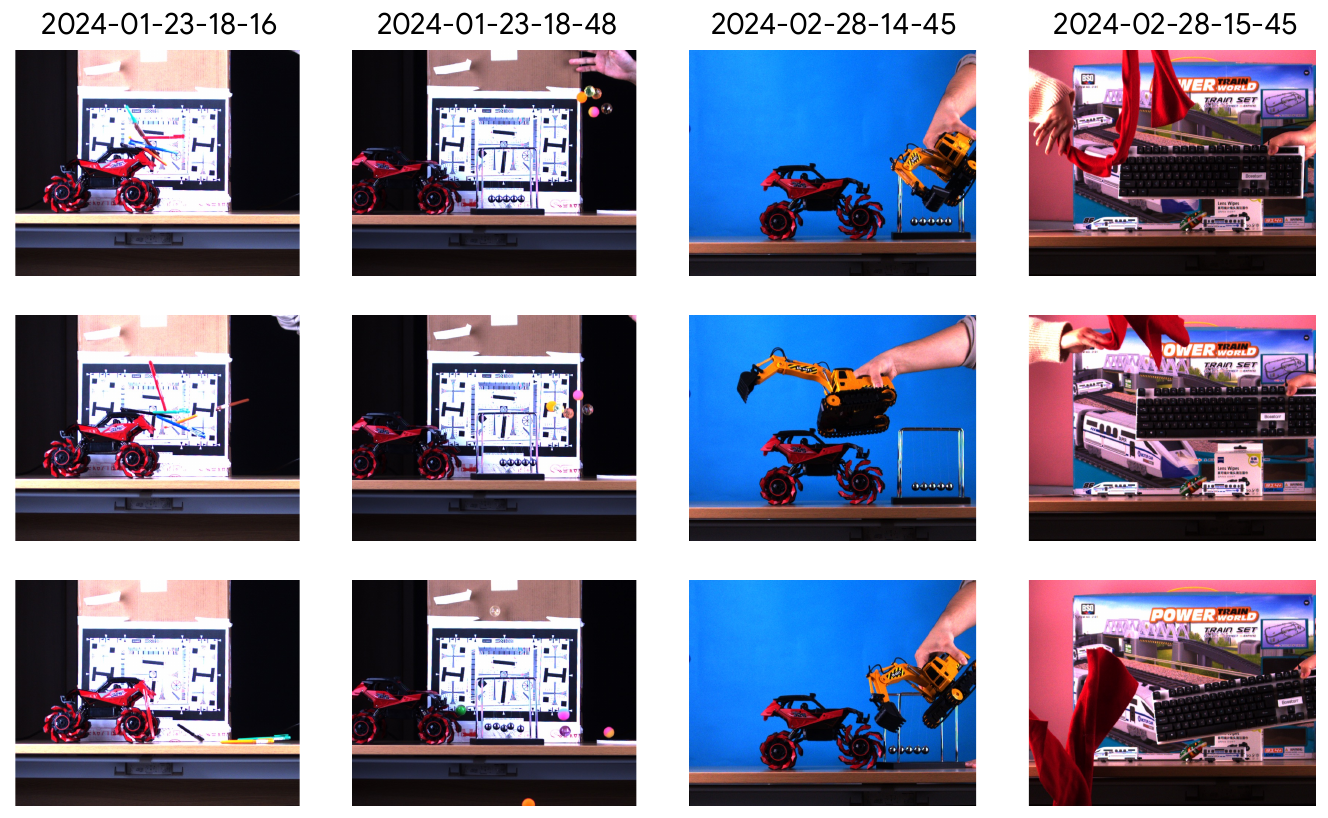}
    \caption{\textbf{Scenes selected for the main quantitative results on HQ-EVFI.} Scenes with highly dynamic motion are chosen to evaluate models under complex real-world motion.}
\label{fig:hq_subsetlist}
\end{figure}

\subsection{Full video frame interpolation result on HQ-EVFI}
\label{sec:supp_hq}

\begin{wraptable}{r}{0.4\linewidth}
\vspace{-35pt}
\centering
\caption{\textbf{Video frame interpolation result on full HQ-EVFI dataset.}}
\label{table:vfi-hqevfi}
\resizebox{1.0\linewidth}{!}{%
\renewcommand{\arraystretch}{1.1}
\begin{tabular}{l|cccc}
\toprule

\multirow{2}{*}{\textbf{Method}}
& \multicolumn{4}{c}{$\times$7} \\

& FID$\downarrow$ & LPIPS$\downarrow$  & PSNR$\uparrow$ & SSIM$\uparrow$ \\
\midrule
TimeLens-XL~\cite{ma2024timelens} &	24.09&0.0569&26.9754& 0.8696\\
CBMNet-Large~\cite{kim2023event} &10.40	&0.0781&\textbf{31.3721}&0.8855\\
\midrule
\hlrow{myblue}
\OursVFI & \textbf{5.73}&\textbf{0.0417}&27.9414&\textbf{0.9069}\\
\bottomrule
\end{tabular}
}
\vspace{-15pt}
\end{wraptable}
Figure~\ref{fig:hq_subsetlist} shows example frames and scene names from the subset used for the main quantitative evaluation.
We present the full zero-shot results on HQ-EVFI on Table~\ref{table:vfi-hqevfi}. The results show that TETO-VFI achieves the best perceptual metrics. We omit VDM-EVFI from the full comparison due to its substantially higher runtime, as the model operates at twice the original resolution during inference as shown in Table~\ref{table:suppl-vfi-cost}. 
\begin{table}
\centering
\caption{\textbf{Inference time comparison between diffusion based frame interpolation models.} Inference time is measured on a single RTX 3090 GPU. TETO is 8.56 times faster than VDM-EVFI.}
\begin{tabular}{l|c}
\toprule
Methods & Inference time (sec) \\
\midrule
VDM-EVFI~\cite{chen2025revdm}  & 2428.00  \\
\hlrow{myblue}
\Ours\  &    283.60 {\scriptsize\textcolor{mygreen}{($\times$ 8.56)}} \\
\bottomrule
\end{tabular}
\label{table:suppl-vfi-cost}
\end{table}

\subsection{Tracking beyond appearance}
\begin{figure}
    \centering
    \includegraphics[width=\linewidth]{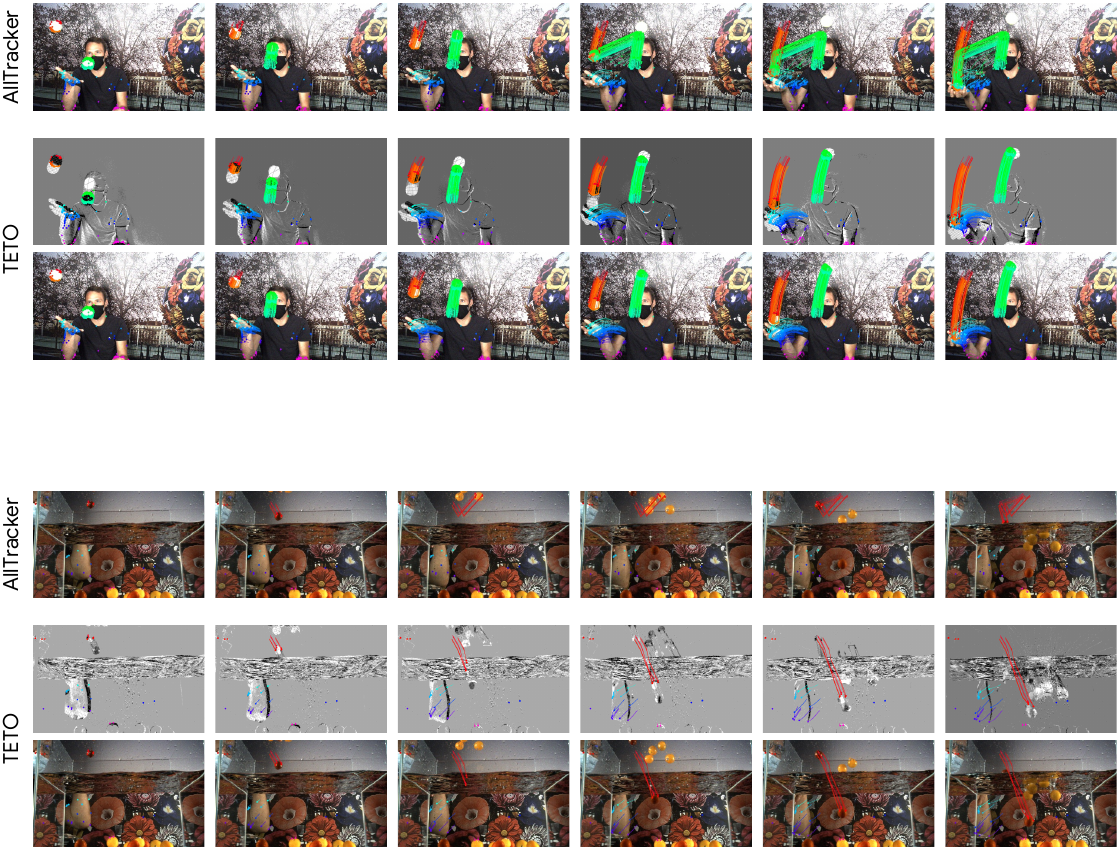}
    \caption{\textbf{Visualization of TETO showing tracking beyond appearance.}}
\label{fig:supp_track_rgb_qual}
\end{figure}

A natural question arising from our distillation framework is whether the student is bounded by the teacher's capability.
Figure~\ref{fig:supp_track_rgb_qual} demonstrates two scenarios where AllTracker, the RGB teacher, fails due to its reliance on appearance, while TETO succeeds by exploiting event-based motion cues.

In Figure~\ref{fig:supp_track_rgb_qual} (Top), two visually similar objects pass close to each other during juggling.
AllTracker confuses their identities at the point of proximity because appearance features alone cannot disambiguate them, causing trajectory swaps.
TETO maintains correct identity throughout because its representation captures continuous temporal motion patterns that distinguish the two objects even when their spatial positions overlap.

In the bottom example, water droplets on a glass surface corrupt the RGB appearance of the object behind it.
AllTracker loses track as the distorted appearance breaks feature matching.
TETO continues to track successfully because event cameras respond to brightness changes regardless of such visual corruption, preserving the underlying motion signal.
Additionally, TETO tracks a partially visible arm seen through the water of an aquarium, where the refracted appearance is too degraded for RGB-based matching but the motion signal remains intact.

These cases illustrate that TETO is not upper-bounded by the teacher's performance.
While pseudo-labels are generated from easy, unoccluded frames where the teacher is reliable, the student learns to associate event temporal structure with motion correspondences rather than imitating the teacher's visual encoding.
This enables generalization to scenarios that the teacher itself cannot handle, consistent with the feature loss ablation in the main paper (Table 7) which shows that encouraging RGB-like intermediate features degrades tracking performance.

\subsection{Qualitative Comparison on point tracking}
\begin{figure*}
    \centering
    \includegraphics[width=\linewidth]{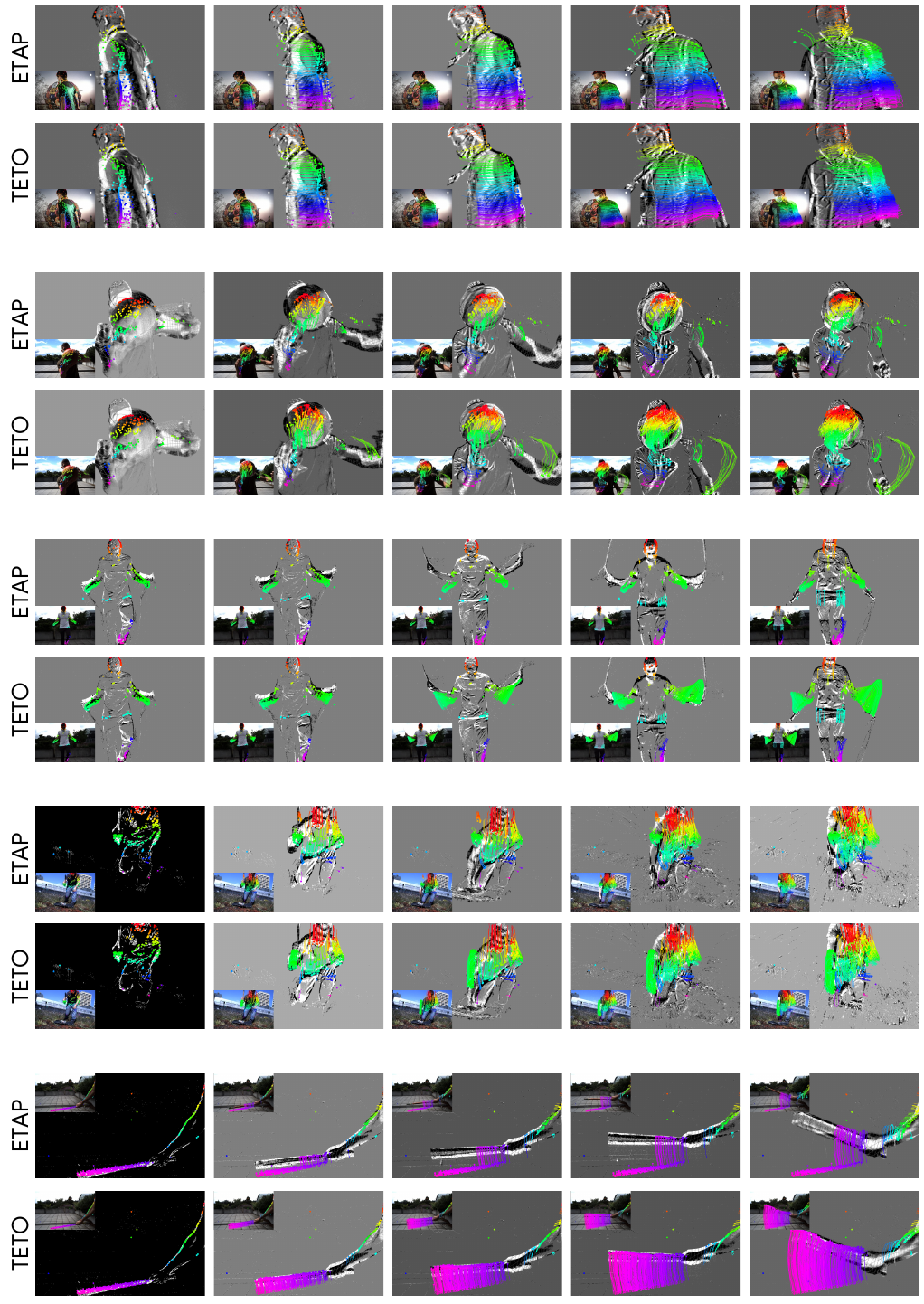}
    \caption{\textbf{Qualitative comparison under dynamic motion.}}
    \label{fig:suppl_bsergb}
\end{figure*}

Additional results of TETO on BS-ERGB are shown in Figure~\ref{fig:suppl_bsergb}.

\begin{figure}
    \centering
    \includegraphics[width=\linewidth]{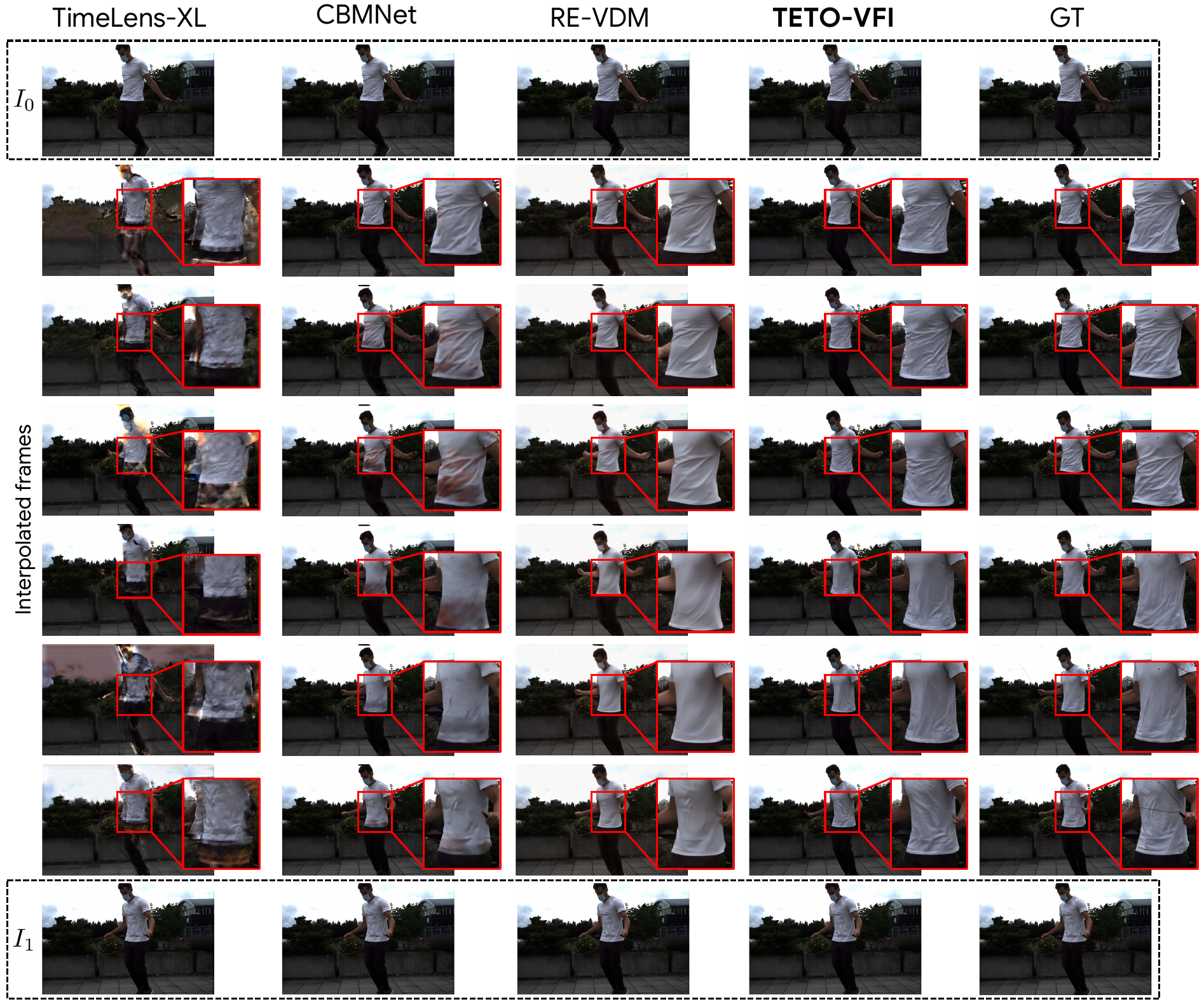}
    \caption{\textbf{Qualitative comparison of video frame interpolation.}}
    \label{fig:supp_vfi_qual1}
\end{figure}

\begin{figure}
    \centering
    \includegraphics[width=\linewidth]{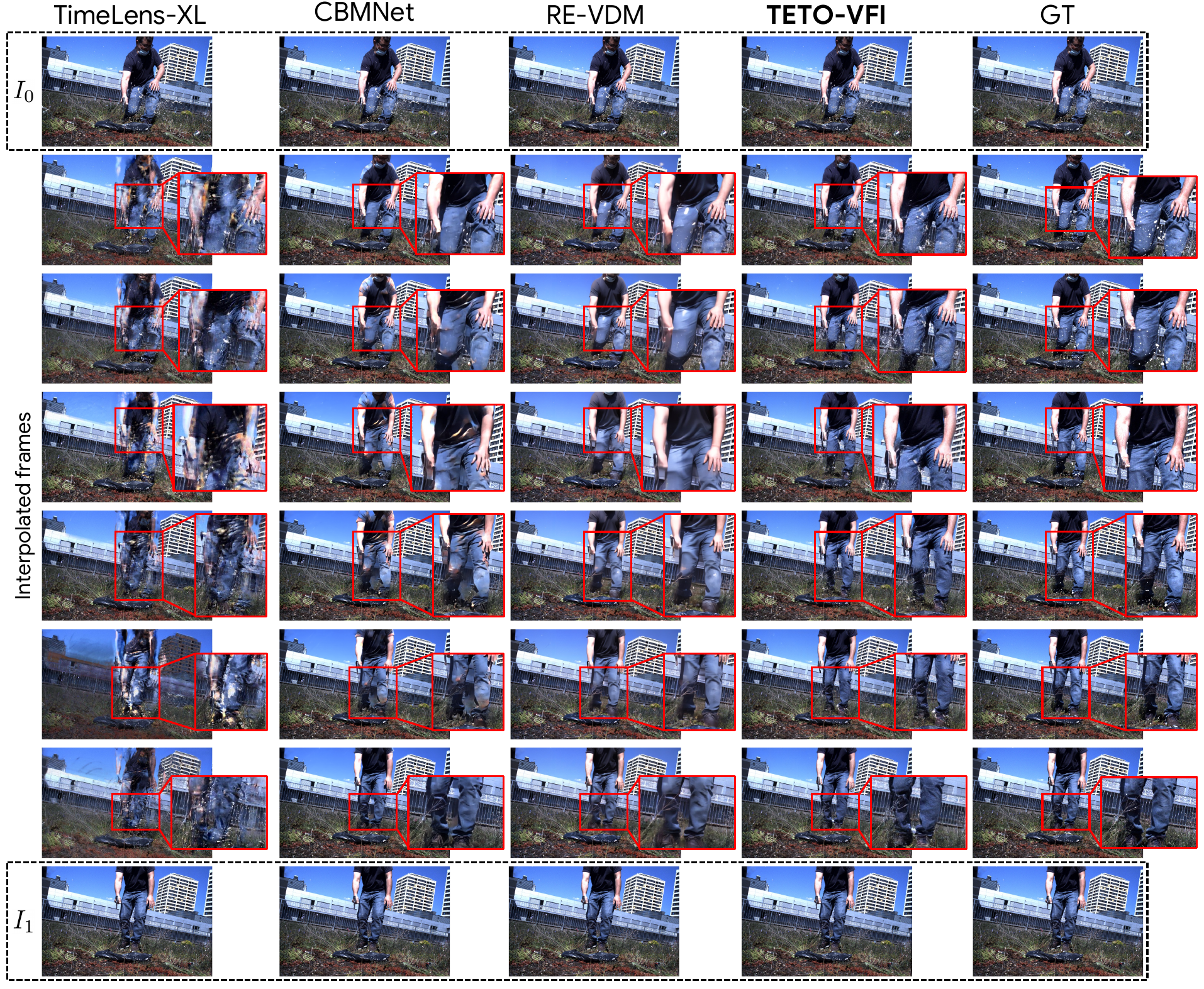}
    \caption{\textbf{Qualitative comparison of video frame interpolation.}}
    \label{fig:supp_vfi_qual2}
\end{figure}

\subsection{Qualitative Comparison on video frame interpolation}

More qualitative comparison of TETO-VFI on BS-ERGB are presented in Figure~\ref{fig:supp_vfi_qual1}, ~\ref{fig:supp_vfi_qual2}. 

\begin{figure}
    \centering
    \includegraphics[width=\linewidth]{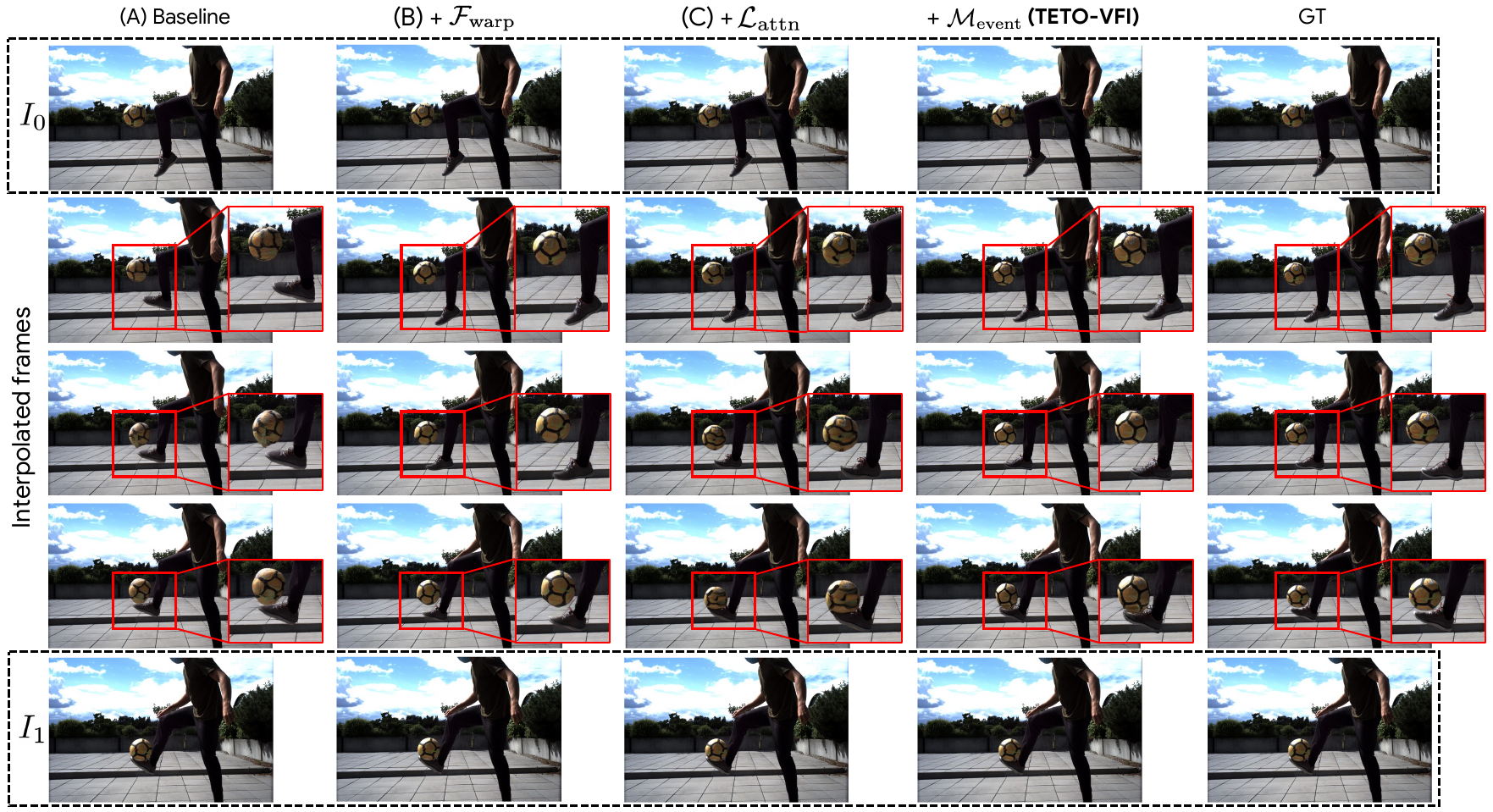}
    \caption{\textbf{Qualitative comparison of video frame interpolation ablations.} (A) is the baseline model without any conditional signal, (B) adds $\mathcal{F}_{\text{warp}}$, and (C) adds $\mathcal{L}_{\text{attn}}$ and adds $\mathcal{M}_{\text{event}}$ for final TETO-VFI. Each result corresponds to the respective entry in the ablation table.}
    \label{fig:VFI_abl_qual}
\end{figure}

\subsection{Qualitative comparison on video frame interpolation ablations.}

Figure~\ref{fig:VFI_abl_qual} shows that the full TETO-VFI model produces the highest-quality interpolation, with sharper details and better motion consistency. Starting from the baseline (A), progressively adding (B) $\mathcal{F}_{\text{warp}}$, (C) $\mathcal{L}_{\text{attn}}$ and finally $\mathcal{M}_{\text{event}}$ improves the interpolation accuracy, while the intermediate variants (A–C) exhibit increasing artifacts and degraded quality.
\section{Inter-Event Interval (IEI) analysis}
\label{sec:supp_iei}

Temporal density is a fundamental characteristic of event camera data, characterizing how frequently brightness changes trigger events over time. To evaluate whether synthetic events faithfully reproduce this temporal behavior, we analyze the temporal gap between real and synthetic events using the Inter-Event Interval (IEI), defined as the time difference between two consecutive events at the same pixel. IEI captures the temporal firing density of event cameras, where smaller values indicate denser temporal information.

We first compute the Inter-Event Interval at each pixel $\mathbf{x}$:
\begin{equation}
\text{IEI}_j^{\mathbf{x}} = t_{j+1}^{\mathbf{x}} - t_j^{\mathbf{x}}, 
\quad j = 1, \dots, N_{\mathbf{x}} - 1,
\end{equation}
where $t_j^{\mathbf{x}}$ denotes the timestamp of the $j$-th event at pixel $\mathbf{x}$ and $N_{\mathbf{x}}$ is the total number of events at that pixel.

We then aggregate all per-pixel IEI values across sequences in the dataset:
\begin{equation}
\text{IEI}_{\text{real}} = \bigcup_{s=1}^{S} \text{IEI}_{\text{real}}^{(s)}, 
\quad
\text{IEI}_{\text{syn}} = \bigcup_{s=1}^{S} \text{IEI}_{\text{syn}}^{(s)},
\end{equation}
where $S$ denotes the number of sequences and $\text{IEI}^{(s)}$ represents the set of per-pixel IEI values from sequence $s$.

To estimate the probability density, we discretize both $\text{IEI}_{\text{real}}$ and $\text{IEI}_{\text{syn}}$ into $B = 200$ uniform bins over a shared range $[0,\, \text{IEI}_{\max}]$ and normalize by the total count and bin width:
\begin{equation}
\hat{f}(b) = \frac{n(b)}{N \cdot \Delta b}, 
\quad b = 1, \dots, B,
\end{equation}
where $n(b)$ is the number of IEI values in bin $b$, $N = |\text{IEI}|$ is the total number of IEI samples, and $\Delta b = \text{IEI}_{\max} / B$ is the bin width. This produces density estimates $\hat{f}_{\text{real}}(\Delta t)$ and $\hat{f}_{\text{syn}}(\Delta t)$ satisfying $\sum_{b=1}^{B} \hat{f}(b)\Delta b = 1$.

Across BS-ERGB, EHPT-XC, and EVRB, real IEI distributions exhibit consistent behavior: a sharp peak at short intervals followed by a rapid decay, with nearly all density concentrated near the origin. This indicates that IEI statistics reflect intrinsic temporal characteristics of event cameras. In contrast, synthetic events generated from frame-based simulators exhibit substantially larger IEI values, with mean intervals approximately $25\times$ larger than those of real data. Because such simulators generate events only at discrete frame interpolation boundaries, they introduce artificial temporal gaps that make synthetic events significantly sparser than real event streams. Consequently, models trained on temporally distorted synthetic data may learn incorrect event timing patterns, leading to degraded performance on real-world event streams.
\section{Limitations}
\label{sec:supp_limitations}

\subsubsection{Fluid motion.}
Fluid motion such as water splashes or flames produces spatially incoherent brightness changes that lack stable correspondences across frames, making reliable point tracking inherently difficult.
This limitation is shared by point tracking methods in general and is not specific to event-based approaches.

\subsubsection{Shadow-induced events.}
Event cameras respond to any brightness change, including those caused by cast shadows.
As a result, TETO may track shadow edges as if they were independent object motion.
Distinguishing shadow-induced events from genuine object motion would require cross-modal reasoning with RGB appearance, which is a natural extension for future work.


\clearpage
\bibliographystyle{splncs04}
\bibliography{main}

@String(ECCV  = {Eur. Conf. Comput. Vis.})

@String(ECCV  = {ECCV})

@inproceedings{lin2022dvs,
  title={Dvs-voltmeter: Stochastic process-based event simulator for dynamic vision sensors},
  author={Lin, Songnan and Ma, Ye and Guo, Zhenhua and Wen, Bihan},
  booktitle={European Conference on Computer Vision},
  pages={578--593},
  year={2022},
  organization={Springer}
}

@article{doersch2022tap,
  title={Tap-vid: A benchmark for tracking any point in a video},
  author={Doersch, Carl and Gupta, Ankush and Markeeva, Larisa and Recasens, Adria and Smaira, Lucas and Aytar, Yusuf and Carreira, Joao and Zisserman, Andrew and Yang, Yi},
  journal={Advances in Neural Information Processing Systems},
  volume={35},
  pages={13610--13626},
  year={2022}
}

@inproceedings{doersch2023tapir,
  title={Tapir: Tracking any point with per-frame initialization and temporal refinement},
  author={Doersch, Carl and Yang, Yi and Vecerik, Mel and Gokay, Dilara and Gupta, Ankush and Aytar, Yusuf and Carreira, Joao and Zisserman, Andrew},
  booktitle={Proceedings of the IEEE/CVF International Conference on Computer Vision},
  pages={10061--10072},
  year={2023}
}

@inproceedings{doersch2024bootstap,
  title={Bootstap: Bootstrapped training for tracking-any-point},
  author={Doersch, Carl and Luc, Pauline and Yang, Yi and Gokay, Dilara and Koppula, Skanda and Gupta, Ankush and Heyward, Joseph and Rocco, Ignacio and Goroshin, Ross and Carreira, Joao and others},
  booktitle={Proceedings of the Asian Conference on Computer Vision},
  pages={3257--3274},
  year={2024}
}

@inproceedings{karaev2024cotracker,
  title={Cotracker: It is better to track together},
  author={Karaev, Nikita and Rocco, Ignacio and Graham, Benjamin and Neverova, Natalia and Vedaldi, Andrea and Rupprecht, Christian},
  booktitle={European conference on computer vision},
  pages={18--35},
  year={2024},
  organization={Springer}
}

@inproceedings{karaev2025cotracker3,
  title={Cotracker3: Simpler and better point tracking by pseudo-labelling real videos},
  author={Karaev, Nikita and Makarov, Yuri and Wang, Jianyuan and Neverova, Natalia and Vedaldi, Andrea and Rupprecht, Christian},
  booktitle={Proceedings of the IEEE/CVF International Conference on Computer Vision},
  pages={6013--6022},
  year={2025}
}

@inproceedings{cho2024local,
  title={Local all-pair correspondence for point tracking},
  author={Cho, Seokju and Huang, Jiahui and Nam, Jisu and An, Honggyu and Kim, Seungryong and Lee, Joon-Young},
  booktitle=ECCV,
  pages={306--325},
  year={2024},
}

@inproceedings{harley2025alltracker,
  title={Alltracker: Efficient dense point tracking at high resolution},
  author={Harley, Adam W and You, Yang and Sun, Xinglong and Zheng, Yang and Raghuraman, Nikhil and Gu, Yunqi and Liang, Sheldon and Chu, Wen-Hsuan and Dave, Achal and You, Suya and others},
  booktitle={Proceedings of the IEEE/CVF International Conference on Computer Vision},
  pages={5253--5262},
  year={2025}
}

@inproceedings{teed2020raft,
  title={Raft: Recurrent all-pairs field transforms for optical flow},
  author={Teed, Zachary and Deng, Jia},
  booktitle={European conference on computer vision},
  pages={402--419},
  year={2020},
  organization={Springer}
}

@article{gallego2020event,
  title={Event-based vision: A survey},
  author={Gallego, Guillermo and Delbr{\"u}ck, Tobi and Orchard, Garrick and Bartolozzi, Chiara and Taba, Brian and Censi, Andrea and Leutenegger, Stefan and Davison, Andrew J and Conradt, J{\"o}rg and Daniilidis, Kostas and others},
  journal={IEEE transactions on pattern analysis and machine intelligence},
  volume={44},
  number={1},
  pages={154--180},
  year={2020},
  publisher={IEEE}
}

@inproceedings{hu2021v2e,
  title={v2e: From video frames to realistic DVS events},
  author={Hu, Yuhuang and Liu, Shih-Chii and Delbruck, Tobi},
  booktitle={Proceedings of the IEEE/CVF conference on computer vision and pattern recognition},
  pages={1312--1321},
  year={2021}
}

@inproceedings{tulyakov2021time,
  title={Time lens: Event-based video frame interpolation},
  author={Tulyakov, Stepan and Gehrig, Daniel and Georgoulis, Stamatios and Erbach, Julius and Gehrig, Mathias and Li, Yuanyou and Scaramuzza, Davide},
  booktitle={Proceedings of the IEEE/CVF conference on computer vision and pattern recognition},
  pages={16155--16164},
  year={2021}
}

@article{hong2024unifying,
  title={Unifying feature and cost aggregation with transformers for semantic and visual correspondence},
  author={Hong, Sunghwan and Cho, Seokju and Kim, Seungryong and Lin, Stephen},
  journal={arXiv preprint arXiv:2403.11120},
  year={2024}
}

@article{an2025c3g,
  title={C3G: Learning Compact 3D Representations with 2K Gaussians},
  author={An, Honggyu and Jung, Jaewoo and Kim, Mungyeom and Hong, Sunghwan and Kim, Chaehyun and Fukuda, Kazumi and Jeon, Minkyeong and Han, Jisang and Narihira, Takuya and Ko, Hyuna and others},
  journal={arXiv preprint arXiv:2512.04021},
  year={2025}
}

@inproceedings{kim2023event,
  title={Event-based video frame interpolation with cross-modal asymmetric bidirectional motion fields},
  author={Kim, Taewoo and Chae, Yujeong and Jang, Hyun-Kurl and Yoon, Kuk-Jin},
  booktitle={Proceedings of the IEEE/CVF Conference on Computer Vision and Pattern Recognition},
  pages={18032--18042},
  year={2023}
}

@inproceedings{cho2024benchmark,
    title={A Benchmark Dataset for Event-Guided Human Pose Estimation and Tracking in Extreme Conditions}, author={Cho, Hoonhee and Kim, Taewoo and Jeong, Yuhwan and Yoon, Kuk-Jin},
    booktitle={The Thirty-eight Conference on Neural Information Processing Systems Datasets and Benchmarks Track},
    year={2024} 
}

@article{burner2022evimo2,
  title={Evimo2: an event camera dataset for motion segmentation, optical flow, structure from motion, and visual inertial odometry in indoor scenes with monocular or stereo algorithms},
  author={Burner, Levi and Mitrokhin, Anton and Ferm{\"u}ller, Cornelia and Aloimonos, Yiannis},
  journal={arXiv preprint arXiv:2205.03467},
  year={2022}
}

@inproceedings{tulyakov2022time,
  title={Time lens++: Event-based frame interpolation with parametric non-linear flow and multi-scale fusion},
  author={Tulyakov, Stepan and Bochicchio, Alfredo and Gehrig, Daniel and Georgoulis, Stamatios and Li, Yuanyou and Scaramuzza, Davide},
  booktitle={Proceedings of the IEEE/CVF Conference on Computer Vision and Pattern Recognition},
  pages={17755--17764},
  year={2022}
}

@inproceedings{rebecq2019events,
  title={Events-to-video: Bringing modern computer vision to event cameras},
  author={Rebecq, Henri and Ranftl, Ren{\'e} and Koltun, Vladlen and Scaramuzza, Davide},
  booktitle={Proceedings of the IEEE/CVF Conference on Computer Vision and Pattern Recognition},
  pages={3857--3866},
  year={2019}
}

@article{carion2025sam,
  title={Sam 3: Segment anything with concepts},
  author={Carion, Nicolas and Gustafson, Laura and Hu, Yuan-Ting and Debnath, Shoubhik and Hu, Ronghang and Suris, Didac and Ryali, Chaitanya and Alwala, Kalyan Vasudev and Khedr, Haitham and Huang, Andrew and others},
  journal={arXiv preprint arXiv:2511.16719},
  year={2025}
}

@inproceedings{hamann2025etap,
  title={ETAP: Event-based Tracking of Any Point},
  author={Hamann, Friedhelm and Gehrig, Daniel and Febryanto, Filbert and Daniilidis, Kostas and Gallego, Guillermo},
  booktitle={Proceedings of the Computer Vision and Pattern Recognition Conference},
  pages={27186--27196},
  year={2025}
}

@inproceedings{shen2025blinktrack,
  title={BlinkTrack: Feature Tracking over 80 FPS via Events and Images},
  author={Shen, Yichen and Li, Yijin and Chen, Shuo and Li, Guanglin and Huang, Zhaoyang and Bao, Hujun and Cui, Zhaopeng and Zhang, Guofeng},
  booktitle={Proceedings of the IEEE/CVF International Conference on Computer Vision},
  pages={9298--9308},
  year={2025}
}

@inproceedings{han2025mate,
  title={MATE: Motion-Augmented Temporal Consistency for Event-based Point Tracking},
  author={Han, Han and Zhai, Wei and Cao, Yang and Li, Bin and Zha, Zheng-jun},
  booktitle={Proceedings of the IEEE/CVF International Conference on Computer Vision},
  pages={8340--8349},
  year={2025}
}

@inproceedings{kim2024cmta,
  title={Cmta: Cross-modal temporal alignment for event-guided video deblurring},
  author={Kim, Taewoo and Cho, Hoonhee and Yoon, Kuk-Jin},
  booktitle={European Conference on Computer Vision},
  pages={1--19},
  year={2024},
  organization={Springer}
}

@article{gehrig2021dsec,
  title={Dsec: A stereo event camera dataset for driving scenarios},
  author={Gehrig, Mathias and Aarents, Willem and Gehrig, Daniel and Scaramuzza, Davide},
  journal={IEEE Robotics and Automation Letters},
  volume={6},
  number={3},
  pages={4947--4954},
  year={2021},
  publisher={IEEE}
}

@inproceedings{vecerik2024robotap,
  title={Robotap: Tracking arbitrary points for few-shot visual imitation},
  author={Vecerik, Mel and Doersch, Carl and Yang, Yi and Davchev, Todor and Aytar, Yusuf and Zhou, Guangyao and Hadsell, Raia and Agapito, Lourdes and Scholz, Jon},
  booktitle={2024 IEEE International Conference on Robotics and Automation (ICRA)},
  pages={5397--5403},
  year={2024},
  organization={IEEE}
}

@inproceedings{luo2021exploring,
  title={Exploring simple 3d multi-object tracking for autonomous driving},
  author={Luo, Chenxu and Yang, Xiaodong and Yuille, Alan},
  booktitle={Proceedings of the IEEE/CVF international conference on computer vision},
  pages={10488--10497},
  year={2021}
}

@inproceedings{zheng2023pointodyssey,
  title={Pointodyssey: A large-scale synthetic dataset for long-term point tracking},
  author={Zheng, Yang and Harley, Adam W and Shen, Bokui and Wetzstein, Gordon and Guibas, Leonidas J},
  booktitle={Proceedings of the IEEE/CVF International Conference on Computer Vision},
  pages={19855--19865},
  year={2023}
}

@inproceedings{greff2022kubric,
  title={Kubric: A scalable dataset generator},
  author={Greff, Klaus and Belletti, Francois and Beyer, Lucas and Doersch, Carl and Du, Yilun and Duckworth, Daniel and Fleet, David J and Gnanapragasam, Dan and Golemo, Florian and Herrmann, Charles and others},
  booktitle={Proceedings of the IEEE/CVF conference on computer vision and pattern recognition},
  pages={3749--3761},
  year={2022}
}

@inproceedings{rebecq2018esim,
  title={Esim: an open event camera simulator},
  author={Rebecq, Henri and Gehrig, Daniel and Scaramuzza, Davide},
  booktitle={Conference on robot learning},
  pages={969--982},
  year={2018},
  organization={PMLR}
}

@inproceedings{reda2022film,
  title={Film: Frame interpolation for large motion},
  author={Reda, Fitsum and Kontkanen, Janne and Tabellion, Eric and Sun, Deqing and Pantofaru, Caroline and Curless, Brian},
  booktitle={European Conference on Computer Vision},
  pages={250--266},
  year={2022},
  organization={Springer}
}

@inproceedings{nam2022stereo,
  title={Stereo depth from events cameras: Concentrate and focus on the future},
  author={Nam, Yeongwoo and Mostafavi, Mohammad and Yoon, Kuk-Jin and Choi, Jonghyun},
  booktitle={Proceedings of the IEEE/CVF conference on computer vision and pattern recognition},
  pages={6114--6123},
  year={2022}
}

@inproceedings{kim2025exploring,
  title={Exploring Temporally-Aware Features for Point Tracking},
  author={Kim, In{\`e}s Hyeonsu and Cho, Seokju and Huang, Jiahui and Yi, Jung and Lee, Joon-Young and Kim, Seungryong},
  booktitle={Proceedings of the Computer Vision and Pattern Recognition Conference},
  pages={1962--1972},
  year={2025}
}

@article{kim2025anthrotap,
  title={AnthroTAP: Learning Point Tracking with Real-World Motion},
  author={Kim, In{\`e}s Hyeonsu and Cho, Seokju and Koo, Jahyeok and Park, Junghyun and Huang, Jiahui and Lee, Honglak and Lee, Joon-Young and Kim, Seungryong},
  journal={arXiv preprint arXiv:2507.06233},
  year={2025}
}

@article{nam2025emergent,
  title={Emergent Temporal Correspondences from Video Diffusion Transformers},
  author={Nam, Jisu and Son, Soowon and Chung, Dahyun and Kim, Jiyoung and Jin, Siyoon and Hur, Junhwa and Kim, Seungryong},
  journal={arXiv preprint arXiv:2506.17220},
  year={2025}
}

@article{son2025repurposing,
  title={Repurposing Video Diffusion Transformers for Robust Point Tracking},
  author={Son, Soowon and An, Honggyu and Kim, Chaehyun and Ko, Hyunah and Nam, Jisu and Chung, Dahyun and Jin, Siyoon and Yi, Jung and Min, Jaewon and Hur, Junhwa and others},
  journal={arXiv preprint arXiv:2512.20606},
  year={2025}
}

@inproceedings{awasthi2025mtevent,
  title={Mtevent: A multi-task event camera dataset for 6d pose estimation and moving object detection},
  author={Awasthi, Shrutarv and Gouda, Anas and Franke, Sven and Rutinowski, J{\'e}r{\^o}me and Hoffmann, Frank and Roidl, Moritz},
  booktitle={Proceedings of the IEEE/CVF Conference on Computer Vision and Pattern Recognition},
  pages={5102--5110},
  year={2025}
}

@inproceedings{chaney2023m3ed,
  title={M3ed: Multi-robot, multi-sensor, multi-environment event dataset},
  author={Chaney, Kenneth and Cladera, Fernando and Wang, Ziyun and Bisulco, Anthony and Hsieh, M Ani and Korpela, Christopher and Kumar, Vijay and Taylor, Camillo J and Daniilidis, Kostas},
  booktitle={Proceedings of the IEEE/CVF conference on computer vision and pattern recognition},
  pages={4016--4023},
  year={2023}
}

@inproceedings{yan2024reli11d,
  title={Reli11d: A comprehensive multimodal human motion dataset and method},
  author={Yan, Ming and Zhang, Yan and Cai, Shuqiang and Fan, Shuqi and Lin, Xincheng and Dai, Yudi and Shen, Siqi and Wen, Chenglu and Xu, Lan and Ma, Yuexin and others},
  booktitle={Proceedings of the IEEE/CVF Conference on Computer Vision and Pattern Recognition},
  pages={2250--2262},
  year={2024}
}

@article{hay2025pose,
  title={E-pose: A large scale event camera dataset for object pose estimation},
  author={Hay, Oussama Abdul and Huang, Xiaoqian and Ayyad, Abdulla and Sherif, Eslam and Almadhoun, Randa and Abdulrahman, Yusra and Seneviratne, Lakmal and Abusafieh, Abdulqader and Zweiri, Yahya},
  journal={Scientific data},
  volume={12},
  number={1},
  pages={245},
  year={2025},
  publisher={Nature Publishing Group UK London}
}

@inproceedings{stoffregen2020hqg,
  title={Reducing the sim-to-real gap for event cameras},
  author={Stoffregen, Timo and Scheerlinck, Cedric and Scaramuzza, Davide and Drummond, Tom and Barnes, Nick and Kleeman, Lindsay and Mahony, Robert},
  booktitle={European Conference on Computer Vision},
  pages={534--549},
  year={2020},
  organization={Springer}
}

@inproceedings{guo2025unsupervised,
  title={Unsupervised joint learning of optical flow and intensity with event cameras},
  author={Guo, Shuang and Hamann, Friedhelm and Gallego, Guillermo},
  booktitle={Proceedings of the IEEE/CVF International Conference on Computer Vision},
  pages={7980--7989},
  year={2025}
}

@inproceedings{ma2024timelens,
  title={Timelens-xl: Real-time event-based video frame interpolation with large motion},
  author={Ma, Yongrui and Guo, Shi and Chen, Yutian and Xue, Tianfan and Gu, Jinwei},
  booktitle={European Conference on Computer Vision},
  pages={178--194},
  year={2024},
  organization={Springer}
}

@article{brebion2021rtef,
  title={Real-time optical flow for vehicular perception with low-and high-resolution event cameras},
  author={Brebion, Vincent and Moreau, Julien and Davoine, Franck},
  journal={IEEE Transactions on Intelligent Transportation Systems},
  volume={23},
  number={9},
  pages={15066--15078},
  year={2021},
  publisher={IEEE}
}

@inproceedings{hamann2024motionpriorcm,
  title={Motion-prior contrast maximization for dense continuous-time motion estimation},
  author={Hamann, Friedhelm and Wang, Ziyun and Asmanis, Ioannis and Chaney, Kenneth and Gallego, Guillermo and Daniilidis, Kostas},
  booktitle={European Conference on Computer Vision},
  pages={18--37},
  year={2024},
  organization={Springer}
}

@inproceedings{paredes2021bteb,
  title={Back to event basics: Self-supervised learning of image reconstruction for event cameras via photometric constancy},
  author={Paredes-Vall{\'e}s, Federico and De Croon, Guido CHE},
  booktitle={Proceedings of the IEEE/CVF Conference on Computer Vision and Pattern Recognition},
  pages={3446--3455},
  year={2021}
}

@inproceedings{paredes2023taming,
  title={Taming contrast maximization for learning sequential, low-latency, event-based optical flow},
  author={Paredes-Vall{\'e}s, Federico and Scheper, Kirk YW and De Wagter, Christophe and De Croon, Guido CHE},
  booktitle={Proceedings of the IEEE/CVF international conference on computer vision},
  pages={9695--9705},
  year={2023}
}

@article{you2024vsasm,
  title={Vector-symbolic architecture for event-based optical flow},
  author={You, Hongzhi and Cao, Yijun and Yuan, Wei and Wang, Fanjun and Qiao, Ning and Li, Yongjie},
  journal={arXiv preprint arXiv:2405.08300},
  year={2024}
}

@inproceedings{zhu2019evflownet,
  title={Unsupervised event-based learning of optical flow, depth, and egomotion},
  author={Zhu, Alex Zihao and Yuan, Liangzhe and Chaney, Kenneth and Daniilidis, Kostas},
  booktitle={Proceedings of the IEEE/CVF conference on computer vision and pattern recognition},
  pages={989--997},
  year={2019}
}

@inproceedings{shiba2022secrets,
  title={Secrets of event-based optical flow},
  author={Shiba, Shintaro and Aoki, Yoshimitsu and Gallego, Guillermo},
  booktitle={European Conference on Computer Vision},
  pages={628--645},
  year={2022},
  organization={Springer}
}

@inproceedings{chen2025revdm,
  title={Repurposing pre-trained video diffusion models for event-based video interpolation},
  author={Chen, Jingxi and Feng, Brandon Y and Cai, Haoming and Wang, Tianfu and Burner, Levi and Yuan, Dehao and Fermuller, Cornelia and Metzler, Christopher A and Aloimonos, Yiannis},
  booktitle={Proceedings of the Computer Vision and Pattern Recognition Conference},
  pages={12456--12466},
  year={2025}
}

@inproceedings{gehrig2021raft,
  title={E-raft: Dense optical flow from event cameras},
  author={Gehrig, Mathias and Millh{\"a}usler, Mario and Gehrig, Daniel and Scaramuzza, Davide},
  booktitle={2021 International Conference on 3D Vision (3DV)},
  pages={197--206},
  year={2021},
  organization={IEEE}
}

@inproceedings{liu2025timetracker,
  title={TimeTracker: Event-based Continuous Point Tracking for Video Frame Interpolation with Non-linear Motion},
  author={Liu, Haoyue and Xu, Jinghan and Chang, Yi and Zhou, Hanyu and Zhao, Haozhi and Wang, Lin and Yan, Luxin},
  booktitle={Proceedings of the Computer Vision and Pattern Recognition Conference},
  pages={17649--17659},
  year={2025}
}

@article{han2025d,
  title={D\^{} 2USt3R: Enhancing 3D Reconstruction with 4D Pointmaps for Dynamic Scenes},
  author={Han, Jisang and An, Honggyu and Jung, Jaewoo and Narihira, Takuya and Seo, Junyoung and Fukuda, Kazumi and Kim, Chaehyun and Hong, Sunghwan and Mitsufuji, Yuki and Kim, Seungryong},
  journal={arXiv preprint arXiv:2504.06264},
  year={2025}
}

@inproceedings{wang2024sea,
  title={Sea-raft: Simple, efficient, accurate raft for optical flow},
  author={Wang, Yihan and Lipson, Lahav and Deng, Jia},
  booktitle={European Conference on Computer Vision},
  pages={36--54},
  year={2024},
  organization={Springer}
}

@inproceedings{huang2022real,
  title={Real-time intermediate flow estimation for video frame interpolation},
  author={Huang, Zhewei and Zhang, Tianyuan and Heng, Wen and Shi, Boxin and Zhou, Shuchang},
  booktitle={European conference on computer vision},
  pages={624--642},
  year={2022},
  organization={Springer}
}

@article{chu2025wan,
  title={Wan-move: Motion-controllable video generation via latent trajectory guidance},
  author={Chu, Ruihang and He, Yefei and Chen, Zhekai and Zhang, Shiwei and Xu, Xiaogang and Xia, Bin and Wang, Dingdong and Yi, Hongwei and Liu, Xihui and Zhao, Hengshuang and others},
  journal={arXiv preprint arXiv:2512.08765},
  year={2025}
}

@article{wang2025ati,
  title={Ati: Any trajectory instruction for controllable video generation},
  author={Wang, Angtian and Huang, Haibin and Fang, Jacob Zhiyuan and Yang, Yiding and Ma, Chongyang},
  journal={arXiv preprint arXiv:2505.22944},
  year={2025}
}

@inproceedings{briedis2025controllable,
  title={Controllable tracking-based video frame interpolation},
  author={Briedis, Karlis Martins and Djelouah, Abdelaziz and Ortiz, Rapha{\"e}l and Gross, Markus and Schroers, Christopher},
  booktitle={Proceedings of the Special Interest Group on Computer Graphics and Interactive Techniques Conference Conference Papers},
  pages={1--11},
  year={2025}
}

@inproceedings{jeong2025track4gen,
  title={Track4gen: Teaching video diffusion models to track points improves video generation},
  author={Jeong, Hyeonho and Huang, Chun-Hao P and Ye, Jong Chul and Mitra, Niloy J and Ceylan, Duygu},
  booktitle={Proceedings of the Computer Vision and Pattern Recognition Conference},
  pages={7276--7287},
  year={2025}
}

@inproceedings{geng2025motion,
  title={Motion prompting: Controlling video generation with motion trajectories},
  author={Geng, Daniel and Herrmann, Charles and Hur, Junhwa and Cole, Forrester and Zhang, Serena and Pfaff, Tobias and Lopez-Guevara, Tatiana and Aytar, Yusuf and Rubinstein, Michael and Sun, Chen and others},
  booktitle={Proceedings of the Computer Vision and Pattern Recognition Conference},
  pages={1--12},
  year={2025}
}

@article{zheng2025eventdiff,
  title={EventDiff: A Unified and Efficient Diffusion Model Framework for Event-based Video Frame Interpolation},
  author={Zheng, Hanle and Han, Xujie and Peng, Zegang and Zhang, Shangbin and Du, Guangxun and Zou, Zhuo and Wang, Xilin and Wu, Jibin and Guo, Hao and Deng, Lei},
  journal={arXiv preprint arXiv:2505.08235},
  year={2025}
}

@inproceedings{liu2023tma,
  title={Tma: Temporal motion aggregation for event-based optical flow},
  author={Liu, Haotian and Chen, Guang and Qu, Sanqing and Zhang, Yanping and Li, Zhijun and Knoll, Alois and Jiang, Changjun},
  booktitle={Proceedings of the IEEE/CVF International Conference on Computer Vision},
  pages={9685--9694},
  year={2023}
}

@article{gehrig2024dense,
  title={Dense continuous-time optical flow from event cameras},
  author={Gehrig, Mathias and Muglikar, Manasi and Scaramuzza, Davide},
  journal={IEEE Transactions on Pattern Analysis and Machine Intelligence},
  volume={46},
  number={7},
  pages={4736--4746},
  year={2024},
  publisher={IEEE}
}

@inproceedings{zhou2025bridge,
  title={Bridge frame and event: Common spatiotemporal fusion for high-dynamic scene optical flow},
  author={Zhou, Hanyu and Wang, Haonan and Liu, Haoyue and Duan, Yuxing and Chang, Yi and Yan, Luxin},
  booktitle={Proceedings of the Computer Vision and Pattern Recognition Conference},
  pages={27904--27913},
  year={2025}
}

@article{wan2025wan,
  title={Wan: Open and advanced large-scale video generative models},
  author={Wan, Team and Wang, Ang and Ai, Baole and Wen, Bin and Mao, Chaojie and Xie, Chen-Wei and Chen, Di and Yu, Feiwu and Zhao, Haiming and Yang, Jianxiao and others},
  journal={arXiv preprint arXiv:2503.20314},
  year={2025}
}

@article{liu2026tapformer,
  title={TAPFormer: Robust Arbitrary Point Tracking via Transient Asynchronous Fusion of Frames and Events},
  author={Liu, Jiaxiong and Tan, Zhen and Zhang, Jinpu and Zhou, Yi and Shen, Hui and Chen, Xieyuanli and Hu, Dewen},
  journal={arXiv preprint arXiv:2603.04989},
  year={2026}
}

@inproceedings{zhang2024sim,
  title={From sim-to-real: Toward general event-based low-light frame interpolation with per-scene optimization},
  author={Zhang, Ziran and Ma, Yongrui and Chen, Yueting and Zhang, Feng and Gu, Jinwei and Xue, Tianfan and Guo, Shi},
  booktitle={SIGGRAPH Asia 2024 Conference Papers},
  pages={1--10},
  year={2024}
}
\end{document}